\documentclass{article}

\PassOptionsToPackage{numbers,compress}{natbib}
\usepackage[main, final]{neurips_2026}

\usepackage[utf8]{inputenc}
\usepackage[T1]{fontenc}
\usepackage{hyperref}
\hypersetup{
    colorlinks=true,
    citecolor=green!50!black,
    linkcolor=red,
    urlcolor=blue,
}
\usepackage{url}
\usepackage{booktabs}
\usepackage{amsfonts}
\usepackage{amsmath}
\usepackage{amssymb}
\usepackage{nicefrac}
\usepackage{microtype}
\usepackage{xcolor}
\usepackage{colortbl}
\usepackage{graphicx}
\usepackage{subcaption}
\usepackage[normalem]{ulem}
\usepackage{bm}
\usepackage{mathtools}
\usepackage{enumitem}
\usepackage{wrapfig}

\newcommand{\R}{\mathbb{R}}
\newcommand{\Splus}{\mathbb{S}^n_+}
\newcommand{\nstar}{n^*}
\DeclareMathOperator{\softplus}{softplus}

\ifx\figurename\undefined \def\figurename{Figure}\fi
\renewcommand{\figurename}{Figure.}
\renewcommand{\paragraph}[1]{\textbf{#1}~~}

\newcommand{\Sect}[1]{Section~\ref{#1}}
\newcommand{\Fig}[1]{Figure~\ref{#1}}
\newcommand{\Tbl}[1]{Table~\ref{#1}}
\newcommand{\Equ}[1]{Eq.~\ref{#1}}
\newcommand{\Apx}[1]{Appendix~\ref{#1}}

\title{Multidimensional Observer Model and Perceptual Dimensions of Human Image Quality Assessment}

\author{%
  Sheng Zhao \\
  Department of Computer Science \\
  University of Rochester \\
  \texttt{szhao32@ur.rochester.edu}
  \And
  Weikai Lin \\
  Department of Computer Science \\
  University of Rochester \\
  \texttt{wlin33@ur.rochester.edu}
  \And
  Yuhao Zhu \\
  Department of Computer Science \\
  University of Rochester \\
  \texttt{yzhu@rochester.edu}
}

\begin{document}

\maketitle

\begin{abstract}
Judging image quality is not only ecologically relevant to everyday humans tasks, but also underpins many machine vision tasks such as image generation.
This paper proposes a framework to understand the inherent perceptual space underlying image quality judgment in humans.
We propose a multi-dimensional observer model that represents images as distributions in a latent perceptual space and that models human judgment as comparing noisy samples.
Being constrained by neural representations in the primate ventral stream and fit to large-scale behavioral data, the model enables analysis of perceptual structure while matching the predictive power of existing metrics.
Using this model, we find that the perceptual spaces needed to account for image quality judgment in humans are extremely low-dimensional compared to the image space even when considering its sparsity.
The exact structure of the space (e.g., dimensionalities, information encoded) varies between low-level and high-level quality judgments, suggesting that, despite a shared retinal encoding in the beginning, humans selectively construct task-dependent perceptual spaces in visual decision making.
The project page and source code are at: \href{https://horizon-lab.org/iqa/}{https://horizon-lab.org/iqa/}.

\end{abstract}

\section{Introduction}

The quality of the visual input is ecologically relevant to everyday human tasks~\citep{gibson1979ecological, wang2025raise, higgins1998vision}.
Judging the quality of an image underpins many machine vision tasks (e.g., image generation~\citep{zeng2026videoargus, zeng2026mira, zhao2026reference}, denoising~\citep{elad2023image}, and rendering~\citep{lin2026lowpowar, lin2025powergs}), and has inspired a long line of perceptual metrics, from PSNR and SSIM~\citep{wang2004image} to deep-feature methods such as LPIPS~\citep{zhang2018unreasonable}, DreamSim~\citep{fu2023dreamsim}, PieAPP~\citep{prashnani2018pieapp}, and DISTS~\citep{ding2020image}.
However, these metrics are primarily optimized for predictive accuracy rather than for probing human perception.
Fundamental questions therefore remain: what perceptual space organizes diverse images, how quality judgments are formed within this space, how many dimensions are required, and whether the structure of the space differs between low- and high-level judgments.

This paper proposes a framework to answer these questions.
The key to our approach is a multi-dimensional observer model that formalizes two key assumptions underlying many IQA datasets.
First, the representation of an image in the perceptual space is noisy, both across observers and within an observer.
Second, in a pairwise comparison, the image that is more distant from the clean, reference image is seen as having a lower quality.
Our observer model, thus, maps each image to a distribution in a multi-dimensional, latent metric space, where distance is proportional to perceptual difference.
The quality of an image is then modeled as the \textit{expected} distance between its latent distribution and that of the reference image.

We enforce that this latent space be a simple affine transformation away from the neural representations found in the primate retinal and ventral stream (in area V1, V2, V4, and Inferior Temporal (IT)~\citep{felleman1991distributed,dicarlo2012does, kubilius2019brain}).
The physiologically meaningful representations permit model interpretation using known properties of visual processing.
We fit the observer model using maximum likelihood, based on behavioral data obtained from standard triplet two-alternative forced choice (2AFC) experiments.
On a diverse set of IQA datasets, including BAPPS~\citep{zhang2018unreasonable}, PieAPP~\citep{ prashnani2018pieapp}, and NIGHTS~\citep{fu2023dreamsim},
our observer model approaches human performance and is on par or better than existing perceptual metrics.

Using this framework, we quantitatively probe the perceptual space of human IQA.
We find that human IQA could be accounted for by extremely low-dimensional perceptual spaces: the dimensionality at which the model performance saturates is much lower than that of the image space, even when considering the sparsity present in natural images.
Interestingly, the exact saturating dimensionality varies based on the task complexity.
On low-level tasks where image quality is dictated by distortion such as blur and compression artifacts~\citep{zhang2018unreasonable, prashnani2018pieapp}, the model performance generally saturates within 10 dimensions, whereas for high-level tasks that require semantic-level similarity processing~\citep{fu2023dreamsim}, the model saturates at a dimensionality an order of magnitude higher.

For low-level tasks, incorporating information from higher-order cortical regions, from V1 and V2 up to V4, produces only a slight performance improvement; including signals from area IT in fact degrades the model’s performance.
In contrast, high-level tasks clearly benefit from information originating in high-level cortical areas, with activity from IT markedly surpassing that from V1.

In summary, this paper makes the following contributions:
\begin{itemize}[leftmargin=1.5em, itemsep=2pt, parsep=0pt]
  \item We introduce a multi-dimensional observer model that represents images as noisy samples in a latent space, permitting characterization of the perceptual structure underlying human IQA.
  \item By constraining the space to align with representations in the primate ventral stream and fitting the model to large-scale behavioral data, we achieve 2AFC prediction performance on par with existing perceptual metrics while offering interpretability.
  \item We show that only a small number of perceptual dimensions, far fewer than those of the image space, can account for human quality judgments, and that humans construct task-dependent perceptual spaces despite sharing common retinal representations across tasks.
\end{itemize}

\section{Related work}
\label{sec:related}

\paragraph{Perceptual Metrics.}
Numerous perceptual metrics and loss functions exist, ranging from classic metrics such as PSNR and SSIM~\citep{wang2004image, wang2003multiscale, zhang2011fsim} to learning-based metrics such as LPIPS~\citep{zhang2018unreasonable}, DreamSim~\citep{fu2023dreamsim}, DISTS~\citep{ding2020image}, MILO~\citep{cogalan2025milo}, TOPIQ-FR~\citep{chen2024topiq}, and PieAPP~\citep{prashnani2018pieapp}, which improve prediction accuracy through better feature encoders, multiscale aggregation, or task-specific calibration.
Complementarily, the Visible Difference Predictor family~\citep{mantiuk2011hdr, mantiuk2021fovvideovdp, mantiuk2024colorvideovdp} grounds its predictions in psychophysical models of the human visual system, incorporating contrast-sensitivity functions, chromatic and temporal channels, and visual masking.
All of these methods, are designed to maximize prediction accuracy; none treats the dimensionality and structure of the underlying perceptual space as quantities to be recovered, which is the subject of this work.

\paragraph{Psychometric Scaling and Dimensions of Perception.}
There is a long history of work on psychometric scaling that relates perceptual experience to (differences in) physical stimuli, from classical formulations such as the Weber–Fechner law~\cite{fechner1860elemente} and Stevens’ power law~\cite{stevens1957psychophysical}, to probabilistic models such as Bradley–Terry (BT)~\cite{bradley1952rank} and Thurstone~\cite{thurstone1927law}, and more recent developments~\cite{vacher2024perceptual,perez2019pairwise,maloney2003maximum}.
However, psychometric scaling typically maps stimuli onto a one-dimensional scale, limiting its ability to capture the structure of the underlying perceptual space, which we focus on.

A few recent studies have examined the dimensionality of the perceptual space underlying object similarity~\cite{hebart2020revealing} and material similarity judgments~\cite{schmidt2025material}.
Beyond focusing on different tasks (IQA in this work), these approaches build on the Bradley–Terry model; we discuss its connection to our observer model in \Sect{sec:discussion}.

\section{Method}
\label{sec:method}

\subsection{Observer Model}
\label{sec:observer}
\label{sec:2afc}

\begin{figure}[t]
\centering
\includegraphics[width=\textwidth]{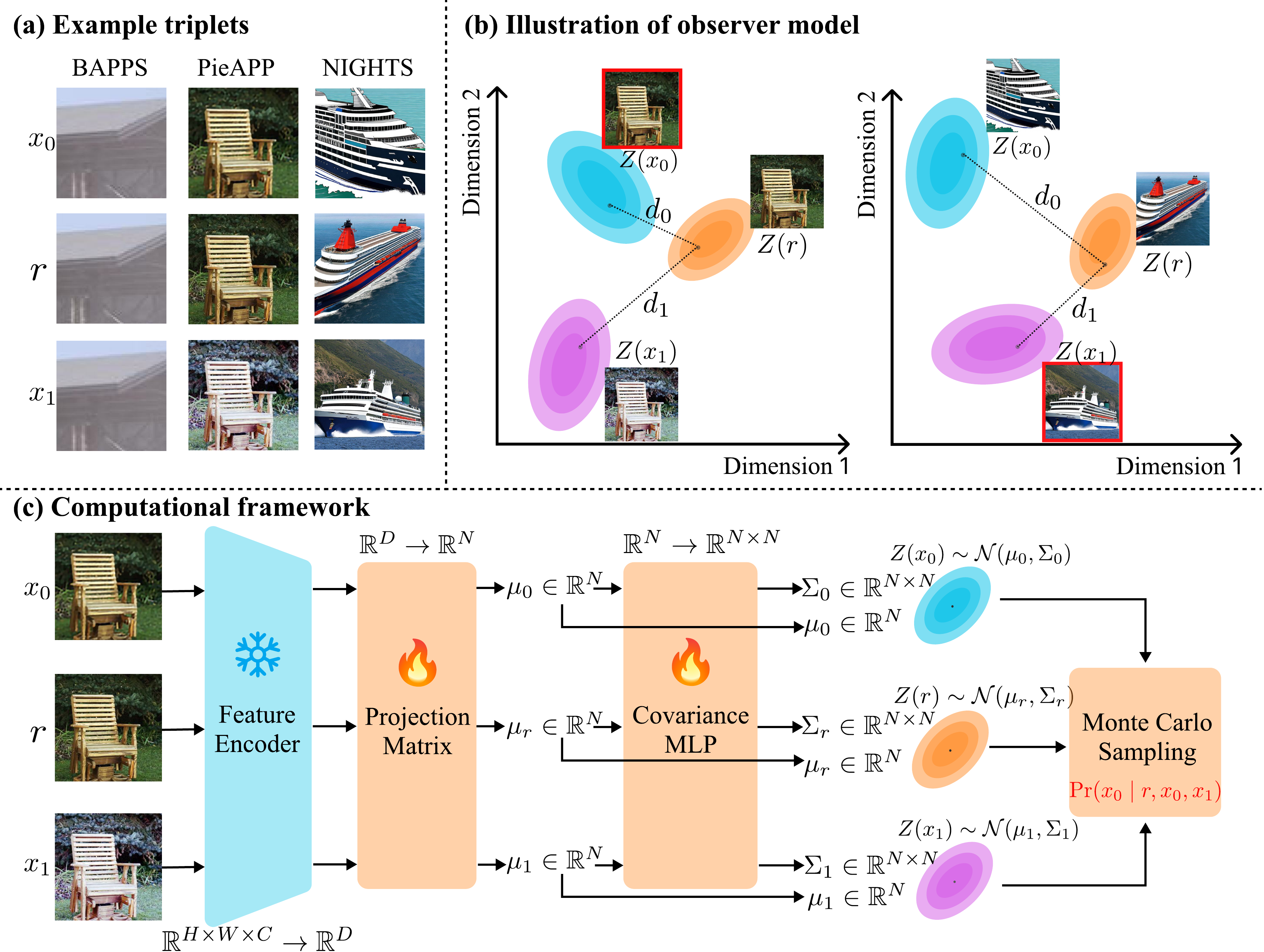}
\caption{\textbf{Overview of the observer model.}
(a): Example triplets (reference $r$ and the two alternatives $x_0$ and $x_1$) from the three 2AFC datasets we evaluate.
(b): Illustration of the observer model.
Each image is modeled as an $N$-dimensional Gaussian random variable $Z(x) \sim \mathcal{N}\!\big(\bm{\mu}(x), \bm{\Sigma}(x)\big)$ in an internal perceptual space.
Each observation amounts of drawing a sample from the three distributions, and the candidate image that is closer to the reference is picked in that trial (visualized with a red bounding box).
(c): The computational framework for the observer model.
}
\label{fig:observer}
\end{figure}

To probe the perceptual space underlying human IQA from behavioral data, we need to make assumptions about how observers respond to images.
These assumptions are formalized in our \textit{observer model}.
Concretely, an observer's visual system responds to each image with a noisy pattern of activity across a population of neurons.
The noise arises both externally (photon shot noise) and internally (neural noise)~\citep{simoncelli200423, bialek1987physical}.
Observers judge image quality by internally comparing these noisy responses across images~\citep{green1966signal}.
Because the internal responses are noisy, the same observer can return different answers on different trials.
Different observers can also disagree on the same trial due to individual differences.
What is reproducible about human behavior is not the individual answer, but its \emph{distribution across trials}.

To formalize these intuitions, we encode each image $x$ in an internal perceptual space as a random variable $Z(x)$, which we assume follows a $N$-variate Gaussian distribution:
\begin{equation}
  \label{eq:observer}
  Z(x) \sim \mathcal{N}\!\big(\bm{\mu}(x),\; \bm{\Sigma}(x)\big),
  \qquad x \in \mathcal{X},
\end{equation}
where $\mathcal{X}$ represents the entire image space, and the functions $\bm{\mu} : \mathcal{X} \mapsto \mathbb{R}^N$ and $\bm{\Sigma} : \mathcal{X} \mapsto \mathbb{R}^{N \times N}$, respectively, map the stimulus $x$ to the mean and covariance matrix of the corresponding distribution in the N-dimensional representational space.
Importantly, the covariance matrix depends on $x$, so different images are permitted to carry different amounts of perceptual ambiguity.
We will describe the parametric form of $\bm{\mu}(x)$ and $\bm{\Sigma}(x)$ in \Sect{sec:fitting}.

How does a human then make a decision?
We use the triplet 2AFC task commonly seen in IQA datasets as an example.
Each trial presents a reference $r$ and two alternatives, $x_0$ and $x_1$.
On any given trial, an observer draws a sample each from the three distributions and obtains $z(r)$, $z(x_0)$, and $z(x_1)$.
We posit that the observer, in that particular trial, regards $x_0$ as having a high quality than $x_1$ if $z(x_0)$ is closer to $z(r)$ than does $z(x_1)$.

We model the distance between $z(x_i)$ and the reference $z(r)$ in the perceptual space using the weighted Minkowski distance, i.e., the $p$-norm:
\begin{equation}
  \label{eq:distance}
  d_i = || z(x_i) -  z(r) ||_p = \bigg(\, \sum_{m=1}^{n} \alpha_m \,\big| z(x_i)_m - z(r)_m \big|^{p} \,\bigg)^{\!1/p},
  \quad i \in \{0, 1\},\;\; \alpha_m \geq 0,\;\; p \geq 1.
\end{equation}
Both the weights $\alpha_m$ and the exponent $p$ are free parameters.
The weights $\alpha_m$ compensate for the different natural magnitudes of the perceptual channels~\citep{nosofsky1986attention}.
The exponent $p$ determines how independently the dimensions can influence the distance, ranging from separable dimensions when $p = 1$ to integral dimensions when $p = 2$~\citep{shepard1964attention, garner1974processing}.
Allowing $p$ to vary let the model adapt to the underlying representational structure (see \Apx{app:learned-p} for experimental results).

On any single trial the observer picks $x_0$ if $d_0 < d_1$, i.e., $\mathbf{1}\{d_0 < d_1\}$, where $\mathbf{1}(\cdot)$ is the indicator function.
Clearly, $d_i$ is a sample from a random variable $D_i = ||Z(x_i) - Z(r)||_p$, where $Z(x_i) - Z(r) \sim \mathcal{N}(\bm{\mu}(x_i) - \bm{\mu}(r), \bm{\Sigma}(x_i) + \bm{\Sigma}(r))$.
Therefore, $\mathbf{1}\{d_0 < d_1\}$ itself is a sample from a Bernoulli random variable, whose expected value is the probability that $x_0$ is picked over $x_1$:
\begin{equation}
  \label{eq:choice}
  \Pr(x_0 \mid r, x_0, x_1) \;=\; \mathbb{E}\!\big[\, \mathbf{1}\{D_0 < D_1\} \,\big].
\end{equation}
While the discussion so far focuses on triplet 2AFC tasks where the reference and both alternatives are simultaneously presented by an observer, our observer model also readily extends to other psychophysical paradigms (e.g., 2AFC tasks where the reference image is not shown~\citep{perez2019pairwise} or subjective rating tasks).
We discuss these extensions in \Apx{app:friqa}.

\subsection{Fitting the Observer Model Using Behavioral Data}
\label{sec:fitting}
\label{sec:model}
\label{sec:mle}

\paragraph{Parameterizing the Mean and Covariance Matrix Field.}
The internal representation of an image $x$ on a given trial depends on its distribution, characterized by its mean $\bm{\mu}(x)$ and covariance matrix $\bm{\Sigma}(x)$ (\Equ{eq:observer}).
Parameterizing these two functions must meet two requirements.
First, $x$ lies in the image space $\mathcal{X}$, which is high dimensional, but the internal representation used for assessing quality may be of a much lower, but unknown, dimensionality.
Therefore, both functions must permit flexible dimensionality reduction without committing to a fixed intrinsic dimensionality.
Second, perceptual responses vary smoothly with stimulus changes, so both $\bm{\mu}(x)$ and $\bm{\Sigma}(x)$ must be smooth over $x$.

A direct projection from the pixel space to $\bm{\mu}(x)$, while meeting the two requirements, is undesirable, as quality assessment is unlikely determined by a weighted sum of pixel values.
Instead, we first encode $x$ as $B(x) \in \mathbb{R}^D$ in a space that is aligned with human visual processing.
For instance, $B(\cdot)$ can be instantiated using pretrained convolutional neural networks such as CORnet-S~\citep{kubilius2019brain} or VGG~\citep{simonyan2015very}.
Features from these networks have been shown to predict neural responses across multiple areas of the visual cortex (e.g., V1, V2, V4, IT) in a layer-wise manner, providing a biologically grounded embedding of images~\citep{schrimpf2018brain, sohn2024explorations, miao2024convolutional}.
Alternatively, one can also represent $B(x)$ using the contrast pyramid of $x$~\citep{peli1990contrast}, which approximates the retinal representation of an image~\citep{mantiuk2024colorvideovdp}.

We then obtain $\bm{\mu}(x)$ via a linear projection:
\begin{equation}
  \label{eq:mean}
  \bm{\mu}(x) \;=\; \mathbf{W}\, B(x) + \mathbf{b},
  \qquad \mathbf{W} \in \R^{N \times D},\quad \mathbf{b} \in \R^N,
\end{equation}
where $N$ is a hyper-parameter that can be tuned.
A linear projection is the minimum mechanism that enables flexibly control of the dimensionality $N$ and preserves the dimensions in $B(x)$ up to an affine transformation, avoiding distortions that would make the latent dimensions harder to interpret.
Both $B(\cdot)$ and the linear projection are smooth over $x$, so $\bm{\mu}(x)$ inherits the property.
We additionally explore using PCA instead of a learned projection matrix for dimensionality reduction; see Appendix~\ref{app:indomain-full} for further details.

The covariance matrix field had an additional constraint that it must be a positive semi-definite matrix for every $x$.
To that end, we rewrite $\bm{\Sigma}(x)$ through the Cholesky decomposition,
\begin{equation}
  \label{eq:chol}
  \bm{\Sigma}(x) \;=\; \mathbf{L}(x)\, \mathbf{L}(x)^{\top},
\end{equation}
and enforce that $\mathbf{L}(x) \in \R^{N \times N}$ is a lower-triangular matrix with non-zero diagonal entries;
this ensures that $\bm{\Sigma}(x)$ is positive semi-definite by construction.
$\mathbf{L}(x)$ is generated by:
\begin{equation}
  \label{eq:mlp}
  \mathbf{L}(x) = s(G_{\psi}(\bm{\mu}(x))),
\end{equation}
where $G_{\psi}$ is a small MLP and $s(\cdot)$ applies a softplus function~\citep{dugas2000incorporating} to the diagonal elements in the MLP output to ensure positivity.
$\bm{\Sigma}(x)$ is smooth over $x$ because the MLP output is naturally smooth over the input $\bm{\mu}(x)$, which is smooth over $x$.
~\Apx{app:architecture} gives the MLP architecture.

\paragraph{Likelihood.}
Given the forms of the mean field $\bm{\mu}(\cdot)$ and the covariance matrix field $\bm{\Sigma}(\cdot)$, we can now describe the (log) likelihood function:
\begin{equation}
  \label{eq:loglik}
  \log p(\mathcal{O} \mid \Theta)
  \;=\; \sum_{i=1}^{T} J^{(i)} \Big[\, y^{(i)} \log P^{(i)} + (1 - y^{(i)}) \log\big(1 - P^{(i)}\big) \,\Big],
\end{equation}
where $\mathcal{O}$ represents observations collected over all the trails (each of which provides whether $x_0$ or $x_1$ is picked),
$P^{(i)} = \Pr\!\big(x_0^{(i)} \mid r^{(i)}, x_0^{(i)}, x_1^{(i)}\big)$ is the probability of choosing $x_0$ on triplet $i$ predicted by our observer model (\Equ{eq:choice}), $y^{(i)}$ is the fraction of actual human observations where $x_0$ is picked on triplet $i$, and $J^{(i)}$ is the total number of observations made on triplet $i$.

The trainable parameter set is $\Theta = \{\mathbf{W}, \mathbf{b}, G_{\psi}, \bm{\alpha}, p\} $, where $\bm{\alpha}$ and exponent $p$ are the per-dimension weights and exponent from \Equ{eq:distance}, $\mathbf{W}$ and $\mathbf{b}$ are used to generate the mean field in \Equ{eq:mean}, and $G_{\psi}$ is the MLP that generates the covariance matrix field in \Equ{eq:chol}.

\paragraph{Model Fitting.}
We estimate the parameters $\Theta$ by maximizing the likelihood on human data provided in 2AFC datasets (\Sect{sec:setup}).
To use gradient descent for optimization, the likelihood function must be differentiable.
The expectation in~\Equ{eq:choice}, however, generally has no closed form~\citep{ennis2014general}.
We resort to a numerical approximation using Monte Carlo (MC) simulations, where we draw $K$ joint samples from the three distributions and averaging the distance comparisons over all $K$ simulation runs:
\begin{equation}
  \label{eq:mc}
  \Pr(x_0 \mid r, x_0, x_1) \;=\; \mathbb{E}\!\big[\, \mathbf{1}\{D_0 < D_1\} \,\big] \approx \frac{1}{K} \sum_{k=1}^{K} (\mathbf{1}\{d_0^{(k)} < d_1^{(k)}\}),
\end{equation}
where $d_i^{(k)}$ is a sample from $D_i$ at MC simulation run $k$.
The indicator function $\mathbf{1}(\cdot)$ is not differentiable.
We use the standard Straight-Through Estimator (STE)~\citep{bengio2013estimating} to allow gradient propagation.
For~\Equ{eq:mc} to be differentiable, the samples themselves must be differentiable.
We achieve this using the standard reparameterization trick~\citep{kingma2014auto}.
In \Apx{app:ste}, we discuss the implementation details of the MC sampling and reparameterization as well as other design alternatives such as replacing the indicator function with a smooth function (e.g., sigmoid).
We also discuss the implications of applying a prior in \Apx{app:wishart}.

\section{Experiment Setup}
\label{sec:setup}

\paragraph{Feature Encoder.}
Our primary choice of encoder $B(\cdot)$ in \Equ{eq:mean} is CORnet-S~\citep{kubilius2019brain}, a convolutional neural network with four major computational blocks.
The output of each block has been shown to align with neural recordings of one of the four areas of the primate ventral stream: V1, V2, V4, and IT.
We spatially pool each channel of each area's output, obtaining one vector per area of dimension $D = 64, 128, 256, 512$ at V1, V2, V4, and IT, respectively.
We additionally concatenate these four vectors into a composite vector of dimension $D = 960$;
we call this encoder ``CORnet-S Multi''.

While CORnet-S encodes images into the cortical space, we also experiment with an encoder that builds a contrast pyramid from an image~\citep{peli1990contrast}.
The contrast pyramid, built from the Laplacian pyramid~\citep{burt1987laplacian}, represents the local contrast at different image regions at different spatial-frequency bands.
It approximates the spatial processing by the retina and LGN~\citep{wandell1995foundations}, and is widely used in perceptual metrics ~\citep{mantiuk2024colorvideovdp, lubin1995visual, ramasubramanian1999perceptually} and perceptual loss functions~\citep{tariq2023perceptually, tursun2019luminance}.

No encoder is fine-tuned and all features are precomputed.
The implementation details are in~\Apx{app:architecture}.
We also evaluate VGG-16~\citep{simonyan2015very} pretrained on ImageNet classification, which has long been used as perceptual features~\citep{gatys2015texture, johnson2016perceptual, zhang2018unreasonable}.
Results on it are shown in \Apx{app:indomain-full}.

\paragraph{Datasets.}
We use three widely used triplet 2AFC datasets spanning both low- and high-level IQA tasks: BAPPS~\citep{zhang2018unreasonable} and PieAPP~\citep{prashnani2018pieapp}, where images within each triplet differ due to low-level distortions (e.g., blur, noise, compression), and NIGHTS~\citep{fu2023dreamsim}, where variation arises from the stochasticity of text-to-image generation rather than explicit distortions.

We also evaluate on four other popular IQA datasets, which either do not use 2AFC tasks (LIVE~\citep{sheikh2006statistical}, KADID-10k~\citep{lin2019kadid}, and CSIQ~\citep{larson2010csiq}) or do not provide raw pairwise comparison data (TID2013~\citep{ponomarenko2015image}).
They do, however, provide a quality score/rating for each image in the form of (differential) Mean Opinion Score (MOS).
We discuss how our method extends to these rating datasets in \Apx{app:datasets}.

\begin{table}[t]
\centering
\caption{Performance of our observer model and comparison with other perceptual metrics.
The baseline metrics are grouped as follows: methods that have the capability to predict choice probabilities in 2AFC tasks (Ours, CVVDP~\citep{mantiuk2024colorvideovdp}, and PieAPP~\citep{prashnani2018pieapp}), methods that predict quality scores in the form of DMOS/MOS (DISTS~\citep{ding2020image}, TOPIQ-FR~\citep{chen2024topiq}, and MILO~\citep{cogalan2025milo}) and in the form of distance in a learned embedding space  (LPIPS~\citep{zhang2018unreasonable}, DreamSim~\citep{fu2023dreamsim}, OpenCLIP~\citep{cherti2023reproducible}, DINOv3~\citep{simeoni2025dinov3}, and CLIP~\citep{radford2021learning}), and classical, parameter-free metrics (SSIM~\citep{wang2004image} and PSNR).
The KL divergence can only be calculated for models that output choice probabilities.
Ceilings are followed by their per-triplet 95\% confidence interval (CI) averaged over all triplets.
\textbf{Bold} = best, \uline{underline} = second \& third best.
}
\label{tab:baseline}
\setlength{\tabcolsep}{4pt}
\resizebox{\textwidth}{!}{%
\begin{tabular}{@{}l >{\columncolor[gray]{0.93}}c >{\columncolor[gray]{0.95}}c cc|ccc|ccccc|cc@{}}
\toprule
Dataset
  & \rotatebox{60}{Ceiling $\pm$ 95\% CI}
  & \rotatebox{60}{CORnet-S-Multi}
  & \rotatebox{60}{CVVDP~\citep{mantiuk2024colorvideovdp}}
  & \rotatebox{60}{PieAPP~\citep{prashnani2018pieapp}}
  & \rotatebox{60}{DISTS~\citep{ding2020image}}
  & \rotatebox{60}{TOPIQ-FR~\citep{chen2024topiq}}
  & \rotatebox{60}{MILO~\citep{cogalan2025milo}}
  & \rotatebox{60}{DreamSim~\citep{fu2023dreamsim}}
  & \rotatebox{60}{OpenCLIP~\citep{cherti2023reproducible}}
  & \rotatebox{60}{DINOv3~\citep{simeoni2025dinov3}}
  & \rotatebox{60}{CLIP~\citep{radford2021learning}}
  & \rotatebox{60}{LPIPS~\citep{zhang2018unreasonable}}
  & \rotatebox{60}{SSIM~\citep{wang2004image}}
  & \rotatebox{60}{PSNR}
  \\
\midrule
BAPPS (Agr\,$\uparrow$) & .797$\pm$.183 & \textbf{.688} & .645 & .629 & \uline{.678} & .667 & .634 & \uline{.683} & .653 & .640 & .624 & .662 & .641 & .633 \\
\hspace{0.8em}cnn    & .885$\pm$.112 & \textbf{.831} & .809 & .765 & .813 & .808 & .800 & \uline{.825} & .800 & .796 & .773 & \uline{.814} & .809 & .801 \\
\hspace{0.8em}color  & .764$\pm$.212 & \textbf{.634} & .610 & .612 & .615 & .616 & .596 & \uline{.627} & .578 & .572 & .560 & .588 & \uline{.624} & .624 \\
\hspace{0.8em}deblur & .740$\pm$.224 & \uline{.596} & \uline{.594} & .547 & \textbf{.599} & .599 & .584 & .591 & .560 & .572 & .539 & .573 & .594 & .590 \\
\hspace{0.8em}f.int  & .760$\pm$.208 & .621 & .590 & .606 & \uline{.626} & .610 & .558 & \textbf{.632} & .611 & .613 & .593 & \uline{.627} & .555 & .543 \\
\hspace{0.8em}s.res  & .801$\pm$.181 & \textbf{.697} & .658 & .619 & \uline{.692} & .686 & .645 & \uline{.695} & .670 & .624 & .627 & .690 & .649 & .642 \\
\hspace{0.8em}trad.  & .858$\pm$.135 & \textbf{.790} & .609 & .703 & \uline{.749} & .695 & .614 & \uline{.771} & .744 & .734 & .716 & .714 & .600 & .573 \\
BAPPS (KL\,$\downarrow$) & -- & \textbf{.170} & .265 & .236 & -- & -- & -- & -- & -- & -- & -- & -- & -- & -- \\
\addlinespace[2pt]
PieAPP (Agr\,$\uparrow$) & .766$\pm$.105 & \uline{.710} & .591 & \uline{.717} & .690 & .651 & .616 & \textbf{.719} & .661 & .651 & .671 & .640 & .584 & .584 \\
PieAPP (KL\,$\downarrow$) & -- & .088 & .985 & \textbf{.083} & -- & -- & -- & -- & -- & -- & -- & -- & -- & -- \\
\addlinespace[2pt]
NIGHTS (Acc\,$\uparrow$) & 1.000$\pm$0 & \uline{.859} & .703 & .629 & \uline{.860} & .633 & .621 & \textbf{.957} & .848 & .851 & .837 & .745 & .621 & .569 \\
\bottomrule
\end{tabular}%
}
\end{table}

\paragraph{Evaluation metrics.}
\label{par: eval_metrics}
For BAPPS and PieAPP, which provide raw pairwise comparison data for each trial (whether $x_0$ or $x_1$ is picked), there are two common ways to evaluate the performance.
First, we report the ``agreement score'' used by BAPPS~\citep{zhang2018unreasonable}:
$\frac{1}{T} \sum_i^T(p_i\, q_i + (1 - p_i)(1 - q_i))$, where $p_i$ is the fraction of trials that, say, $x_0$ is picked in triplet $i$, $q_i$ is the predicted probability that $x_0$ would be picked, given by \Equ{eq:mc}, and $T$ is the total number of triplets.
This metric measures the probability that a model and an average human observer would agree on every trial.
The theoretical ceiling is $\frac{1}{T} \sum_i^T\max\{p_i, 1-p_i\}$, which is 79.7\% and 76.5\% for BAPPS and PieAPP, respectively.
The inter-observer agreement, i.e., the expectation that two independent human observers agree on every trial, is $\frac{1}{T}\sum_i^T(p_i^2 + (1 - p_i)^2)$; this value is 73.1\% on BAPPS and 68.7\% on PieAPP.

Unlike our observer model, many existing perceptual metrics predict a quality score for an image (relative to a reference) instead of directly modeling the choice probabilities in a 2AFC task.
When using these metrics to model the choice probabilities in a 2AFC task, a common method is to predict the image having a higher predicted score to always be picked~\citep{prashnani2018pieapp}.
For a fair comparison against these metrics, we also binarize the output of our observer model through majority voting of the predicted probabilities.
Effectively, $q_i \in \{0, 1\}$.
A sufficiently good model could out-perform the inter-observer agreement, but should always be worse than the theoretical ceiling.

There exist perceptual metrics that could model choice probabilities in a 2AFC task, as they produce perceptual scores via psychometric scaling that are consistent with the observed choice probabilities~\citep{mantiuk2024colorvideovdp, perez2019pairwise, prashnani2018pieapp}.
For comparison with these metrics, we compute and report the KL divergence between the choice distribution of human observers and that of the model predictions.

For NIGHTS whose labels are binary to begin with, we binarize the output of our observer model through majority voting; the agreement score then reduces to binary error rate (BER)~\citep{prashnani2018pieapp}.
We report 1 - BER as the prediction accuracy.

\paragraph{Protocol.}
For each feature encoder and dataset pair, we sweep the perceptual dimension $N$.
As $N$ increases, the prediction accuracy gradually saturates, at which point we have an empirical estimate of a task's \text{intrinsic perceptual dimensionality}.
We define the saturation point $n^*$ as the smallest $N$ at which the model is able to explain $95\%$ of model's own above-chance performance: $\nstar = \min\{ N: \frac{m(N) - 0.5}{m^* - 0.5} > 0.95 \}$, where 0.5 is the chance level of 2AFC tasks, $m(N)$ is the model's performance at dimension $N$, and $m^*$ is the maximum performance when $N$ is unconstrained.
We use the following hyperparameters: Adam with learning rate $10^{-3}$ and cosine annealing, batch size $256$, $50$ epochs, and $K = 64$ Monte Carlo samples.

\section{Results}
\label{sec:results}

\subsection{Our Observer Model Predict Human 2AFC Tasks Accurately}
\label{sec:baseline}

\begin{table}[t]
\centering
\begin{minipage}[t]{0.46\textwidth}
\centering
\caption{Saturation point $\nstar$ of our observer model based on the agreement score.
The results still hold true under KL divergence, whose results and full per-distortion breakdown are shown in \Apx{app:indomain-full}.
}
\label{tab:main}
\scriptsize
\setlength{\tabcolsep}{2pt}
\begin{tabular}{@{}l cc cc cc@{}}
\toprule
& \multicolumn{2}{c}{BAPPS} & \multicolumn{2}{c}{PieAPP} & \multicolumn{2}{c}{NIGHTS} \\
\cmidrule(lr){2-3} \cmidrule(lr){4-5} \cmidrule(lr){6-7}
Feature Encoder & Agr & $\nstar$ & Agr & $\nstar$ & Acc & $\nstar$ \\
\midrule
Contrast Pyramid & .645 & 3 & .647 & 2 & .703 & 7 \\
CORnet-S V1    & .685 & 4 & .701 & 6 & .747 & 8 \\
CORnet-S V2    & .689 & 3 & .712 & 5 & .793 & 14 \\
CORnet-S V4    & .688 & 4 & .714 & 8 & .855 & 48 \\
CORnet-S IT    & .677 & 6 & .702 & 14 & .853 & 96 \\
CORnet-S Multi & .688 & 3 & .710 & 9 & .859 & 96 \\
\bottomrule
\end{tabular}
\end{minipage}
\hfill
\begin{minipage}[t]{0.52\textwidth}
\centering
\caption{Agreement score comparison on BAPPS between frequency-space sparsity (dictated by top $K$ dimensions/ch. retained after DCT) vs. perceptual-space sparsity (dictated by $N$).
Shaded: a pair of comparison where the sparsity is the same (K=N/3).}
\label{tab:sparsity-ablation}
\scriptsize
\setlength{\tabcolsep}{2pt}
\begin{tabular}{@{}l c >{\columncolor[gray]{0.88}}c >{\columncolor[gray]{0.88}}c ccc@{}}
\toprule
& \multicolumn{2}{c}{Perceptual Space} & \multicolumn{4}{c}{Frequency Space (top-$K$/ch.)} \\
\cmidrule(lr){2-3} \cmidrule(lr){4-7}
Feature Encoder & $N{=}2$ & $N{=}6$ & $K{=}2$ & $K{=}10$ & $K{=}20$ & $K{=}200$ \\
\midrule
Contrast Pyramid & .634 & .637 & .536 & .574 & .586 & .627 \\
CORnet-S V1    & .666 & .682 & .552 & .589 & .615 & .674 \\
CORnet-S V2    & .672 & .685 & .563 & .608 & .632 & .686 \\
CORnet-S V4    & .667 & .685 & .564 & .615 & .636 & .686 \\
CORnet-S IT    & .647 & .669 & .564 & .609 & .628 & .674 \\
CORnet-S Multi & .670 & .686 & .566 & .612 & .635 & .684 \\
\bottomrule
\end{tabular}
\end{minipage}
\end{table}

Our goal is \textit{not} to set a new state of the art in predicting human 2AFC choices, but to probe the human perceptual space of quality judgment.
Nevertheless, to serve as a reliable tool for this purpose, our method must achieve competitive performance relative to existing metrics, as we demonstrate below.

\Tbl{tab:baseline} shows our accuracy on the three datasets and that of other perceptual metrics.
See the caption for the descriptions of the baseline metrics.
In this experiment we use the CORnet-S-Multi as the feature encoder.
On the agreement score metric, our model reaches 0.688 agreement on BAPPS, outperforming all baselines; on PieAPP, our model narrowly trails the best method and is within 0.056 of the theoretical ceiling; on NIGHTS, our model essentially ties with DISTS and falls only behind DreamSim, which uses modern foundation models.
On KL divergence, our observer model performs the best compared to CVVDP and PieAPP on the BAPPS dataset, and narrowly trails as the second best on the PieAPP dataset.

In general, our model (and other metrics) performs worse on BAPPS than on PieAPP.
One reason is that BAPPS collects only 5 observations per triplet, whereas PieAPP collects up to 40.
A lower sample count means higher degrees of uncertainty and wider confidence intervals with respect to the underlying population distribution, as shown in \Tbl{tab:baseline}.

\subsection{Extremely Low-Dimensional Perceptual Spaces Account for Human Judgment}
\label{sec:sparsity}

As the dimensionality of the perceptual space increases, the predictive performance gradually saturates.
We show that quality judgments are well described by an extremely low-dimensional perceptual space, and the low dimensionality is not a trivial consequence of the sparsity of natural images.

\Fig{fig:saturation} shows the performance our observer model on the three datasets as a function of $N$.
We show six variants of our model, each using a different feature encoder.
\Tbl{tab:main} shows the saturating $N$ and the corresponding performance.
Overall, the saturating dimensionalities $n^*$ are extremely low.
On both low-level datasets, the saturating dimensionality is generally around 5, and on the high-level dataset, the dimensionality is about $10\times$ higher, but generally the saturating dimensionalities are much lower than the dimensionality of the image space (i.e., height $\times$ weight $\times$ color channels).

A low saturating dimensionality could potentially be trivially explained by the well-known sparsity (compressibility) of natural images in the frequency domain.
To test this, we construct a frequency-space control experiment.
Inspired by how JPEG exploits image sparsity for compression, for every image color channel (in the YUV color space) we apply the Discrete Cosine Transform (DCT) and retain only the top-$K$ DCT coefficients.
We then reconstruct the image through an inverse DCT, and apply our observer model to reconstructed images (see \Apx{app:sparsity-control} for details).
Since truncation is done per channel, an image would have the same sparsity in the truncated frequency space and in our learned perceptual space when $K=N/3$, where $N$ is the dimensionality of the perceptual space.

\Tbl{tab:sparsity-ablation} shows the performance of our observer model and the DCT control under varying $N$s and $K$s.
Matching the performance of our $N{=}6$ model requires roughly two orders of magnitude more dimensionalities in the DCT control.
The shaded columns share the same sparsity ($N=6, K=2$), in which case the DCT control is about 0.1 lower in the agreement score.
Thus, the low dimensionality we observe cannot simply be attributed to pixel-space sparsity.

\subsection{Task Complexity Shapes Perceptual Dimensions and Representational Requirements}
\label{sec:tasks}

\begin{figure}[t]
\centering
\includegraphics[width=\textwidth]{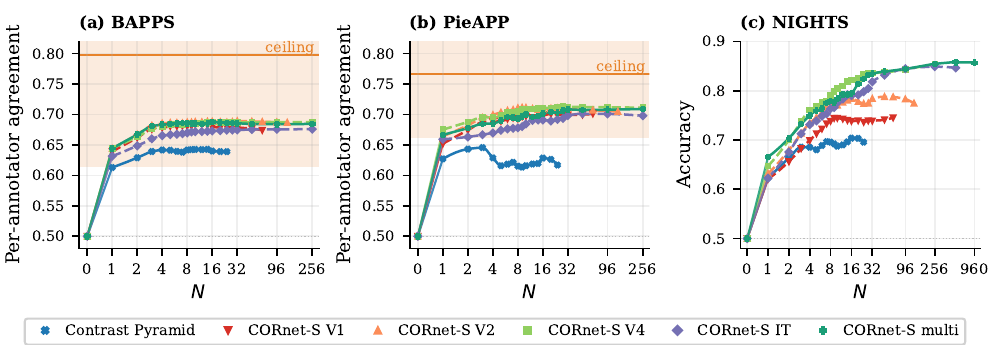}
\caption{Prediction performance of our observer model under different $N$.
We show six variants of our model, each using a different feature encoder.
Low-level IQA datasets (BAPPS, PieAPP) plateau within a handful of dimensions while the high-level IQA dataset NIGHTS requires more dimensionalities.
For BAPPS and PieAPP, we show the 95\% CI along with the theoretical ceiling derived the sample distribution; the $y$-axes are capped for clarity.
NIGHTS ceiling is 1 and CI is 0.
}
\vspace{-10pt}
\label{fig:saturation}
\end{figure}

The dimensionality of perceptual judgment is strongly task-dependent, and the difference is consistent across encoder choices.
As shown in \Fig{fig:saturation} and \Tbl{tab:main}, for BAPPS and PieAPP, where judgments are driven by low-level image distortions~\citep{zhang2018unreasonable, prashnani2018pieapp}, saturation occurs at $\nstar = 3$--$8$;
for NIGHTS, where images differ in mid-level and semantic properties~\citep{fu2023dreamsim}, $\nstar$ grows much larger, reaching $96$ at IT.
This is consistent with the intuition that tasks in NIGHTS involve judging higher-level, semantic differences, which in turn demand substantially higher-dimensional representations.
The same holds on another high-level dataset, Deep-Ranking~\citep{wang2014learning} (\Apx{app:deep-ranking}).

In both tasks across the three datasets, encoding images with the contrast pyramid gives the lowest performance;
this indicates that retinal representation alone can not fully account for human judgments\footnote{In fact, when we enforce the use of the contrast pyramid and increase the number of trainable parameters via $N$, the model starts fitting the noise, leading to a drop in prediction performance as seen in the results.}.
Low-level IQA judgments depend on low-level, local image statistics, and are best supported by early-to-intermediate visual areas (V1 to V4), which all have a retinotopic map of the visual field~\cite{felleman1991distributed}.
Moving all the way to IT, which is primarily responsible for object recognition and does not have a clear retinotopic mapping, actually reduces alignment with human choices.
In contrast, high-level judgments depend instead on structural and semantic organization, so they benefit from later visual representations such as IT, while performance using V1 and V2 representations are markedly lower.

We also find that a model trained on BAPPS incurs only a negligible drop in accuracy when applied directly to PieAPP, another low-level dataset, relative to a model trained on PieAPP itself.
In contrast, performance degrades substantially when the same BAPPS-trained model is applied to NIGHTS, a high-level IQA dataset, compared to a model trained on NIGHTS.
This corroborates that the perceptual space structure varies with task complexity.
\Apx{app:cross-results} provides the detailed results.

\subsection{Interpreting the Perceptual Dimensions}
\label{sec:interpret}

\begin{figure}[t]
\centering
\includegraphics[width=\textwidth]{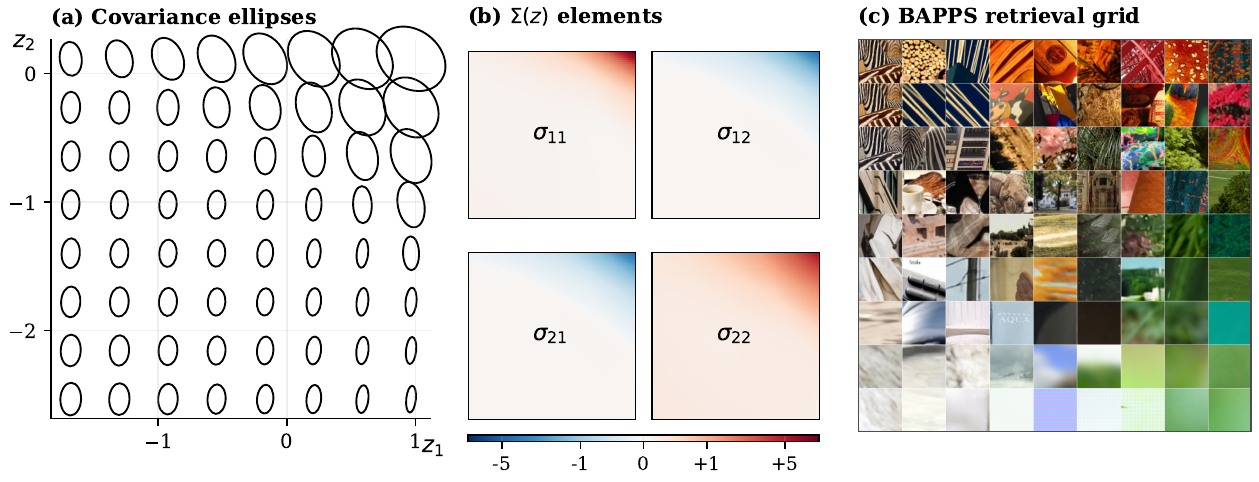}
\caption{Interpretation of the learned $N = 2$ model (CORnet-S-Multi, BAPPS).
(a) The smoothly varying covariance matrix field over the two learned dimensions $z_1$ and $z_2$ sampled in a 8$\times$8 grid; each sampled covariance contour is visualized as an ellipse (log-scale) bounded by two standard deviations.
(b) A complementary visualization of the covariance matrix field showing how $\sigma^2_{z1}$, $\sigma^2_{(z1, z2)}$, $\sigma^2_{(z2, z1)}$, and $\sigma^2_{z2}$ vary individually in the learned 2D space.
(c) $9 \times 9$ images that, when mapped to the perceptual space, vary vertically along the $z_1$ dimension from low to high and vary horizontally along the $z_2$ dimension from low to high.
Reading along these two axes reveals that $z_1$ captures texture and $z_2$ captures color.
More results are in \Apx{app:perceptual-interp}.}
\vspace{-5pt}
\label{fig:interpret}
\end{figure}

If the human judgment can be captured in a relatively low-dimensional perceptual spaces, a natural question is: what does each dimension encode?
Using the CORnet multi-layer model trained on BAPPS, \Fig{fig:interpret} shows the results at $N{=}2$.
\Fig{fig:interpret}(a) visualizes the covariance matrix field over the two learned dimensions, $z_1$ and $z_2$, sampled on a 8 $\times$ 8 grid.
\Fig{fig:interpret}(b) offers a complementary view, visualizing how each element of the covariance matrix --- $\sigma^2_{z_1}$, $\sigma_{(z_1, z_2)}$, $\sigma_{(z_2, z_1)}$, and $\sigma^2_{z_2}$ --- varies over the 2D perceptual space.
The covariance matrix field varies smoothly, reflecting the fact that human responses change gradually with the input stimulus.

\Fig{fig:interpret}(c) visualizes the visual information encoded by each dimension using the BAPPS dataset.
We do that by comparing images that share a similar value in one dimension but differ significantly along the other dimension:
images in the same row (column) share a similar $z_1$ ($z_2$) value but differ in $z_2$ ($z_1$).
Ostensibly, $z_1$ encodes the texture/spatial frequency information while $z_2$ encodes the color information.
These directions closely match classical distinctions in early vision, although they are learned here purely from behavioral data without any prior on their semantics.
Results on the other two datasets are in \Apx{app:perceptual-interp}, and the general conclusions hold.

As $N$ increases, the semantic meaning of each dimension becomes more fine-grained and harder to describe in natural language.
We show in \Apx{app:perceptual-interp} that: 1) perceptual dimensions are still orthogonal to each other in high-dimensional perceptual spaces, and 2) perceptual dimensions for high-level IQA tasks encode more object semantics while those for low-level IQA tasks encode more color and texture information.

\section{Discussion}
\label{sec:discussion}

\paragraph{Connection to Bradley-Terry Model.}
The Bradley-Terry (BT) model is a classical psychometric scaling method commonly used to predict choice distribution pairwise comparisons~\cite{bradley1952rank}, and has been applied to triplet 2AFC tasks~\cite{prashnani2018pieapp, hebart2020revealing, schmidt2025material,perez2019pairwise}.
The BT model assumes that the perceived quality score of an image follows the Gumbel distribution~\cite{tsukida2011analyze}.
In \Apx{app:bt}, we show the BT model and our method give similar predictive power, and our method can be thought of as giving a low-level interpretation of the phenomenological BT model.

\paragraph{Limitations.}
Since we jointly learn the mean and the covariance matrix field, the choice probability of a triplet remains the same if one scales the mean and covariance of each image proportionally.
Thus, it is difficult to compare distributions between images not in the same triplet.
For the same reason, as in prior observer models~\citep{hong2025comprehensive, nejatbakhsh2023estimating}, the learned covariance should be read as the model's internal noise rather than a direct measure of human uncertainty.
In the future, we can gather more direct observations on the mean field by, for instance, collecting neural recordings during IQA tasks or integrating cross-content judgments.

\section{Conclusion}

This paper proposes a framework to model image quality judgment in humans.
We show that the perceptual space in which humans assess image quality is extremely low dimensional, much lower than that of the image space even when considering the sparsity of the image space.
We also show that humans construct task-dependent perceptual spaces from the shared retinal representations.
By providing a principled bridge between behavioral data and the (learned) perceptual space, we open a new window into probing human perceptual processing of images.


{
\small
\bibliographystyle{unsrtnat}
\bibliography{references}
}

\newpage
\appendix

\section{Architecture and Methodology Details}
\label{app:architecture}

\paragraph{Feature Encoders.}
We evaluate three frozen encoder families, each spatially pooled to a manageable feature dimension $D$ before entering the observer head.
\textbf{CORnet-S}~\citep{kubilius2019brain} provides four ventral-stream-aligned areas (V1, V2, V4, IT) with channel counts $64, 128, 256, 512$; each area is global-average-pooled (GAP) to a vector, and CORnet-S-Multi concatenates the four GAP vectors ($D = 960$).
\textbf{VGG-16}~\citep{simonyan2015very} (ImageNet-pretrained) supplies post-block-3 and post-block-5 features, each GAP-pooled (\textbf{VGG-conv3}, $D=256$; \textbf{VGG-conv5}, $D=512$); their LPIPS-style five-block GAP concatenation~\citep{zhang2018unreasonable} forms \textbf{VGG-multi} ($D=1{,}472$); we additionally retain $4{\times}4$ and $2{\times}2$ spatial grids on conv3 and conv5 to test whether pooling is the bottleneck (\textbf{VGG-conv3 $(4{\times}4)$}, $D=4{,}096$; \textbf{VGG-conv5 $(2{\times}2)$}, $D=2{,}048$).
\textbf{Contrast Pyramid}~\citep{peli1990contrast, burt1987laplacian} calculates the contrast pyramid of an image using Laplacian pyramid.
We use the implementation in ColorVideoVDP~\citep{mantiuk2024colorvideovdp} and stops before the CSF weighting step;
we GAP-pool within each (channel, band) combination ($D=24$).
No encoder is fine-tuned; per-dataset features are computed once and cached on disk for reuse across the full $N$-sweep.

\paragraph{Covariance Matrix Field MLP.}
We parameterize the per-image covariance $\bm{\Sigma}(x)$ of Eq.~\eqref{eq:observer} via its Cholesky factor to guarantee positive definiteness:
\begin{equation*}
  \bm{\Sigma}(x) \;=\; \mathbf{L}(x)\,\mathbf{L}(x)^{\!\top} \;\in\; \Splus,
  \qquad \mathbf{L}(x) \in \R^{N \times N}_{\text{lower-tri}}.
\end{equation*}
The $N(N{+}1)/2$ entries of $\mathbf{L}(x)$ are produced by a small MLP $G_\psi$,
\begin{equation*}
  G_\psi: \R^{N} \xrightarrow{\text{Linear}(N, 64)} \text{ReLU} \xrightarrow{\text{Linear}(64, 64)} \text{ReLU} \xrightarrow{\text{Linear}(64, \nicefrac{N(N+1)}{2})} \R^{N(N+1)/2},
\end{equation*}
with $\softplus(\cdot) + \varepsilon$ ($\varepsilon = 10^{-4}$) applied to the diagonal entries of $\mathbf{L}(x)$ so that each diagonal is strictly positive and $\bm{\Sigma}(x)$ is non-degenerate.
The input to $G_\psi$ is the projected mean $\bm{\mu}(x) \in \R^N$.

\paragraph{MC Sampling and Reparameterization.}
\label{app:reparameterization}
For~\Equ{eq:mc} to be differentiable, the samples themselves must be differentiable.
We achieve this using the standard reparameterization trick~\citep{kingma2014auto}
instead of sampling $z(x)$ directly, we draw standard normal noise $\bm{\varepsilon}$ and push it through $\mathbf{L}(x)$.

In particular, on every trial, the model estimates $\Pr(x_0 \mid r, x_0, x_1)$ of~\Equ{eq:mc} by the following sequence.
We first compute the three means $\bm{\mu}(r), \bm{\mu}(x_0), \bm{\mu}(x_1)$ and the three Cholesky factors $\mathbf{L}(r), \mathbf{L}(x_0), \mathbf{L}(x_1)$ once per triplet, then sample neural responses $K$ times through MC simulation:
\begin{equation}
  \label{eq:reparam}
  z(x)^{(k)} \;=\; \bm{\mu}(x) \;+\; \mathbf{L}(x)\, \bm{\varepsilon}^{(k)},
  \qquad \bm{\varepsilon}^{(k)} \sim \mathcal{N}(\mathbf{0}, \mathbf{I}_n).
\end{equation}

For each MC index $k = 1, \ldots, K$, we draw three standard Gaussians $\bm{\varepsilon}_r^{(k)}, \bm{\varepsilon}_0^{(k)}, \bm{\varepsilon}_1^{(k)} \sim \mathcal{N}(\mathbf{0}, \mathbf{I}_n)$, form the two candidate--reference differences $z(x_0)^{(k)} - z(r)^{(k)}$ and $z(x_1)^{(k)} - z(r)^{(k)}$ via ~\Equ{eq:reparam}, compute the weighted $p$-norms $d_0^{(k)}, d_1^{(k)}$ from~\Equ{eq:distance}, and record the indicator $\mathbf{1}\{d_0^{(k)} < d_1^{(k)}\}$.
Averaging these $K$ simulations yields an estimate of $\Pr(x_0 \mid r, x_0, x_1)$.

\paragraph{STE.}
\label{app:ste}
The indicator $\mathbf{1}\{d_0^{(k)} < d_1^{(k)}\}$ in \Equ{eq:mc} is not differentiable, so optimizing $\Theta$ by gradient descent requires a gradient estimator.
We use the popular STE method to allow gradient flow~\cite{bengio2013estimating}.
An alternative is to replace the hard indicator function with a smooth proxy, the sigmoid function being a popular choice in the literature: $\sigma\!\big(\tau\,(D_1 - D_0)\big)$, where $\sigma(\cdot)$ is the sigmoid function and $\tau$ is a learnable temperature controlling how close the smooth function is to the indicator function.

Empirically, we find that STE and the sigmoid method give similar results.
We choose the STE method since with the sigmoid, the covariance matrix field becomes harder to interpret.
This is because the sigmoid function, being smooth, directly gives a probability measure even with just a single sample each from the three distributions ($Z(x_0)$, $Z(x_1)$, and $Z(r)$).
Therefore, the network would have no incentive to learn the covariance matrix field and rely on learning only the mean field (with a very narrow, spatially uniform covariance matrix field).

\paragraph{Parameter Count.}
Our pipeline factorizes into a \emph{frozen} feature encoder and a small \emph{trainable} head; only the head is updated during training, while the backbone is loaded from public pretrained weights and never modified.
\Tbl{tab:app-params} reports trainable / frozen / total parameter counts for our model and the baseline methods used in the the main text.
At its earliest saturation point ($N{=}3$, beyond which validation accuracy on BAPPS no longer improves) and at the BAPPS peak of the Multi-layer backbone reported in \Tbl{tab:main} ($N{=}12$), our trainable head is between one and four orders of magnitude smaller than every other learnable metric in the table, and never larger than $25$K parameters.

\begin{table}[h]
\centering
\small
\caption{Parameter counts comparison.}
\label{tab:app-params}
\begin{tabular}{@{}lrrr@{}}
\toprule
Method & Trainable & Frozen & Total \\
\midrule
DISTS                                            & 2{,}950          & 14.7M (VGG-16)              & 14.7M \\
MILO                                             & 45K              & ---                         & 45K \\
TOPIQ-FR (CFANet, ResNet-50 end-to-end)          & 36.0M            & 53K (Normalization)      & 36.0M \\
DreamSim (LoRA $r{=}16$ on 3$\times$ViT-B/16)    & 1.77M            & 264.2M                      & 266.0M \\
PieAPP                              & 68.4M            & ---                         & 68.4M \\
\midrule
CORnet-S-Multi ($N{=}3$)                  & \textbf{7.7K}    & 52.9M      & 52.9M \\
CORnet-S-Multi ($N{=}12$)                 & \textbf{21.6K}   & 52.9M     & 52.9M \\
\bottomrule
\end{tabular}
\end{table}

\paragraph{DCT Sparsity Control.}
\label{app:sparsity-control}
The control reported in \Sect{sec:sparsity} asks whether the small saturated dimensionality $\nstar$ of our observer simply reflects the well-known compressibility of natural images: if a top-$K$-coefficient reconstruction in some natural image basis already matches the observer's agreement at the matched budget $K = N/3$, then the low $\nstar$ is a property of the image sparsity, not of perceptual processing.

Concretely, for each input image, we first transform it to the YUV color space, apply the DCT, and then keep only the top $K$ coefficients in each color channel.
We then apply the inverse DCT to reconstruct the compressed image.
The reconstructed image is then fed to the same feature encoder as in the main observer model, and the resulting features are passed to a learnable but \text{square} $D \times D$ projection (i.e., no dimensionality reduction) followed by the same covariance predictor as in the main model.
We sweep $K \in \{2, 10, 20, 200\}$ per color channel.
For a $256 \times 256$ image, each channel has $H \cdot W = 65{,}536$ real DCT coefficients, so the swept budgets correspond to retention ratios from $0.003\%$ to $0.3\%$.
The results are shown in \Tbl{tab:sparsity-ablation}.

\paragraph{Datasets.}
\label{app:datasets}
We main evaluate on three triplet 2AFC datasets (BAPPS, NIGHTS, PieAPP) (\Sect{sec:setup}).
We also evaluate on four other popular low-level IQA datasets, which either do not use 2AFC tasks (LIVE Release 2~\citep{sheikh2006statistical}, KADID-10k~\citep{lin2019kadid}, and CSIQ~\citep{larson2010csiq}) or do not provide raw pairwise comparison data (TID2013~\citep{ponomarenko2015image})\footnote{\citet{mikhailiuk2018psychometric} extended TID2013 with additional 2AFC experiments and provide raw pairwise comparison results~\citep{perez2019pairwise, perez2017practical}, but the reference images were not shown during the 2AFC tasks.}.
They do, however, provide a quality score for each image in the form of Mean Opinion Score (MOS) or Differential MOS (DMOS).
We will show in \Apx{app:friqa} how our observer model can be extended to modeling rating-based datasets.
Here, we describe the details of these datasets pertaining to our study.
All datasets are used as released without filtering or relabeling.

\textit{BAPPS}~\citep{zhang2018unreasonable} is a low-level 2AFC triplet dataset of $64 \times 64$ image patches.
Each trial shows a reference (sampled from MIT-Adobe~5K~\citep{fivek2011} for training and RAISE1K~\citep{dang2015raise} for validation) and two corrupted candidates, and the annotator picks the one perceptually closer to the reference.
Distortions span two families: traditional image-processing artifacts (photometric shifts, additive noise, Gaussian and bilateral blur, spatial warps, JPEG) and CNN artifacts produced by randomly parameterized autoencoders trained on denoising / colorization / super-resolution.
Labels were collected with $2$ annotators per triplet on the $151$K-triplet training split and $5$ annotators per triplet on the $36$K-triplet validation split.
The validation set is further stratified into six distortion subsets (traditional, CNN, superresolution, video deblurring, frame interpolation, colorization), which we report separately.
The raw choice distribution is reported for each triplet.

\textit{NIGHTS}~\citep{fu2023dreamsim} is a high-level 2AFC triplet dataset of $768 \times 768$ images synthesized by Stable Diffusion v1.4 from prompts of the form ``an image of a $\langle$category$\rangle$''.
Within a triplet the reference and two candidates share the same prompt but use different random seeds, so variation arises from the stochasticity of the diffusion sampling process rather than any explicit low-level distortion.
Up to ten annotators rated each triplet, yielding a single binary label per triplet through majority vote; the released split contains $15{,}941$ training, $1{,}958$ validation, and $2{,}120$ test triplets ($\approx 20$K total).

\textit{PieAPP}~\citep{prashnani2018pieapp} is a low-level 2AFC triplet dataset built from $200$ Waterloo Exploration reference images~\citep{ma2016waterloo} at $256 \times 256$, each corrupted by a diverse set of image-processing operations and algorithmic artifacts.
For each pair of distorted variants of a given reference, approximately $40$ annotators judge which is perceptually closer to the reference;
the raw choice distribution for each triplet is reported.
The official release provides $77{,}280$ train$+$validation pairs and a $4{,}200$-pair test.

\textit{TID2013}~\citep{ponomarenko2015image} is an FR-IQA dataset with $25$ reference images at $512 \times 384$ and $3{,}000$ distorted images obtained by applying $24$ distortion types at $5$ intensity levels per reference.
On each trial an observer sees the reference together with two distorted versions and selects the one that differs less from the reference; $971$ observers contributed $524{,}340$ such selections.
The average votes that an image receives is used as its perceptual score.

\textit{KADID-10k}~\citep{lin2019kadid} is an FR-IQA dataset with $81$ references at $512 \times 384$ and $10{,}125$ distorted images obtained from $25$ distortion types at $5$ intensity levels per reference.
For each distorted image, the reference and the distorted image are shown side by side and an observer rates the distorted image on a five-point scale from $1$ (``very annoying'') to $5$ (``imperceptible'').
The points are averaged across $30$ ratings per image to yield a score for the image.

\textit{CSIQ}~\citep{larson2010csiq} is an FR-IQA dataset with $30$ references at $512 \times 512$ and $866$ distorted images covering six distortion types (JPEG, JPEG-2000, global contrast decrement, additive pink Gaussian noise, additive white Gaussian noise, Gaussian blur) at $4$--$5$ levels each.
$35$ observers arranged images along a horizontal axis so that physical distances reflect perceived quality differences, producing roughly $5{,}000$ ratings that were converted into perceptual scores.

\textit{LIVE Release~2}~\citep{sheikh2006statistical} is an FR-IQA dataset with $29$ references (typically $768 \times 512$) and $779$ distorted images covering five distortion types: JPEG2000 compression, JPEG compression, additive white noise on RGB, Gaussian blur, and JPEG2000 transmission errors over a fast-fading Rayleigh channel.
Each distorted image is rated on a continuous scale (with each session run by $20$--$29$ subjects and approximately $23$ ratings per image on average) and averaged over subjects to a perceptual score.

\section{Perceptual Space Interpretation and Visualization}
\label{app:perceptual-interp}

\subsection{Perceptual Space Interpretation at N=2 on PieAPP and NIGHTS}
\label{app:perceptual-grids}

\begin{figure}[t]
  \centering
  \begin{minipage}{0.48\linewidth}
    \centering
    \includegraphics[width=\linewidth]{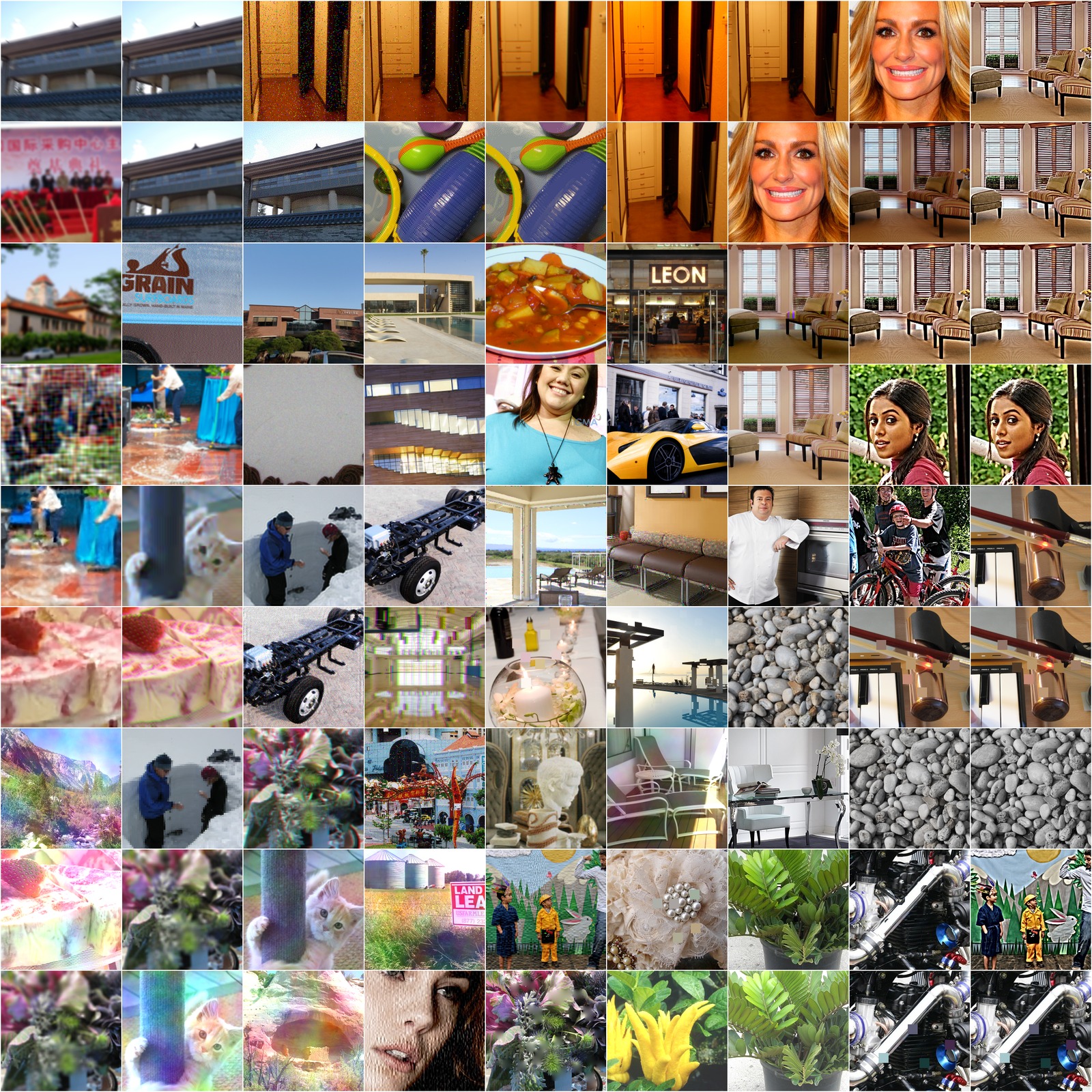}\\
    {\small (b) PieAPP, $N=2$.}
  \end{minipage}\hfill
  \begin{minipage}{0.48\linewidth}
    \centering
    \includegraphics[width=\linewidth]{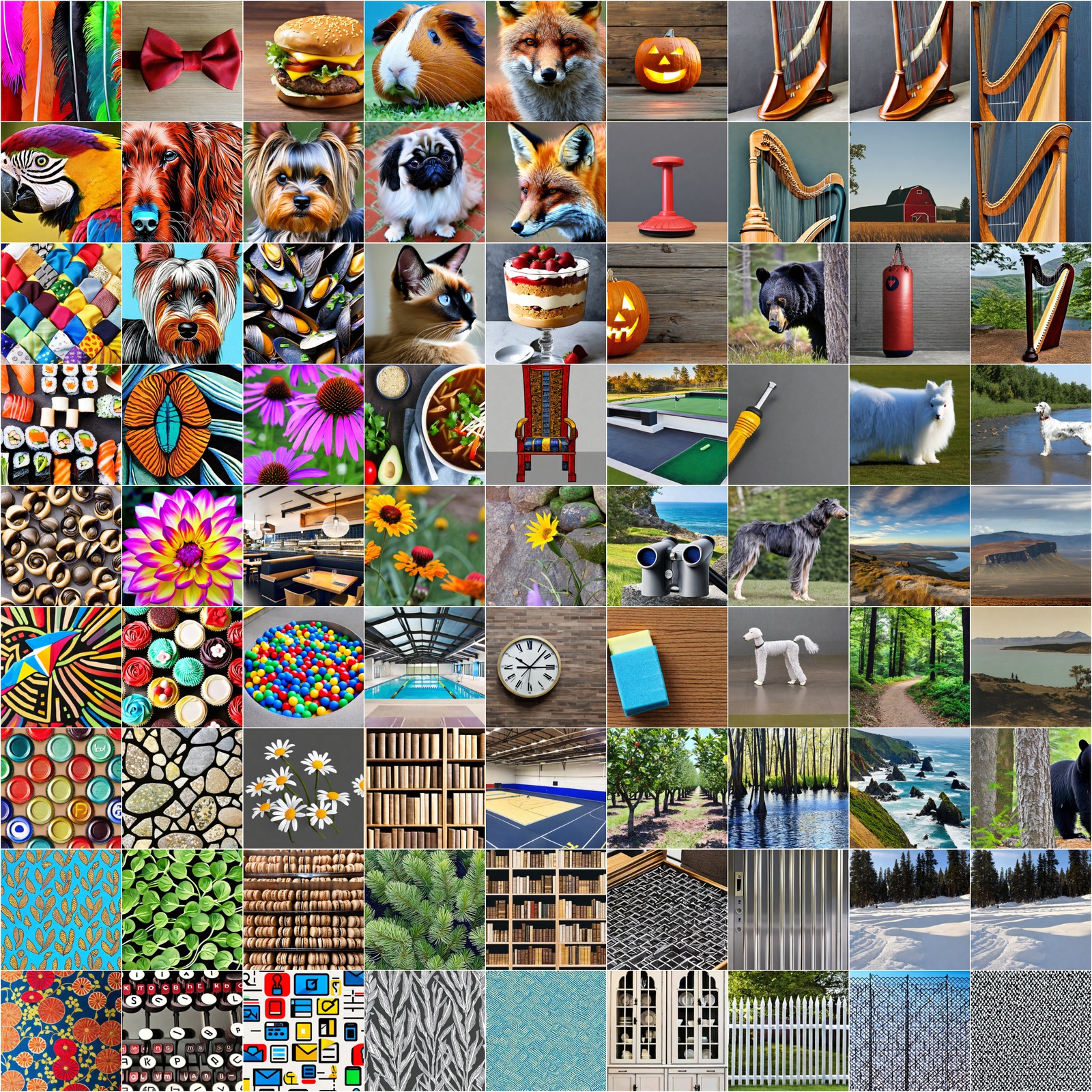}\\
    {\small (c) NIGHTS, $N=2$.}
  \end{minipage}
  \caption{Visualizing the learned $N{=}2$ perceptual space under CORnet-S Multi across datasets.
  Each cell of the $9\times 9$ grid is an image from the dataset situated near that point in the learned space, so traversing horizontally and vertically shows how the model's representation morphs as one of the two latent dimensions varies low to high.
  Across the three datasets, the two axes consistently expose a texture/structure direction crossed with a color (PieAPP) or scene-content (NIGHTS) direction.
  See \Fig{fig:interpret}(c) for results on BAPPS.}
  \label{fig:appendix-grids}
\end{figure}

\Fig{fig:appendix-grids} shows nearest-neighbour retrieval grids at $N=2$ for PieAPP and NIGHTS.
Both datasets recover dimensions similar in spirit to those in BAPPS (\Fig{fig:interpret}(c)).
On PieAPP, the vertical axis ranges from large-scale repetitive structure (architectural corridors, doorways) at the top to fine cluttered texture (gravel, mechanical parts) at the bottom, and the horizontal axis ranges from warm, colorful outdoor scenes to predominantly muted indoor objects, similar in spirit to BAPPS's texture and color axes.
On NIGHTS, one vertical axis tracks scene multiplicity, with single foreground subjects (a fox, a hamster, a harp) at the top and multiple repeating elements (fences, radiator stripes) at the bottom;
the horizontal axis tracks color, with more colored images on the left and near-monochromatic images on the right.

\subsection{Perceptual Space Interpretation at High Dimensionalities}
\label{app:dim-encoding}

When the dimensionality increases, interpreting the dimensions becomes more difficult, since each dimension carries more fine-grained meaning that might not even be easily described in natural language.
We use a different method.
Since our perceptual space $\mathbb{R}^N$ is an affine transformation from the space given by the feature encoder $\mathbb{R}^D$, each perceptual dimension is a weighted sum of the feature dimensions.
We first characterize the concept extracted by each feature dimension using the classic work BRODEN by \citet{bau2017network}.
From the weights in the learned projection matrix, we can then aggregate the meaning associated with each dimension in the perceptual space.

\begin{figure}[t]
  \centering
  \includegraphics[width=\linewidth]{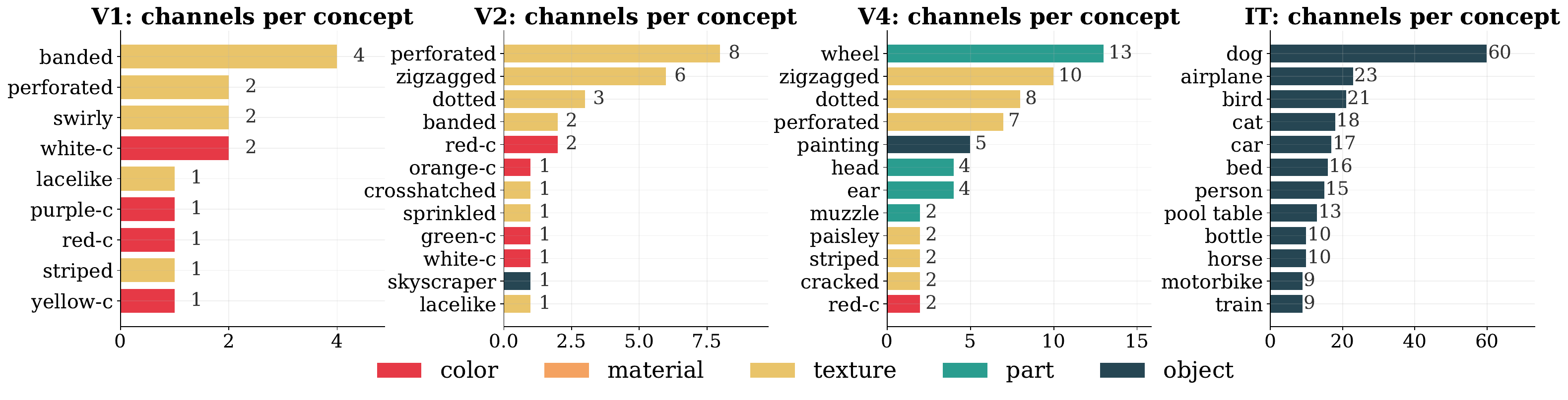}
  \caption{BRODEN~\citep{bau2017network} concept distribution per CORnet-S layer at the ratio $\geq 5\times$ alignment threshold.
  Top: fraction of channels whose top concept lies in each pixel-level category.
  Bottom: per-layer top concepts ranked by the number of channels claiming them.}
  \label{fig:appendix-layer}
\end{figure}

\paragraph{Interpreting Each CORnet-S Channel.}
BRODEN classifies the visual concept extracted by each feature into five main categories each with a few detailed sub-categories associated with it: color, material, texture, part, and object.
We find the top-3 sub-categories associated with each CORnet-S channel.
Each subplot in \Fig{fig:appendix-layer} shows the number of CORnet-S channels associated with each sub-category (see implementation details later).
We see that V1 channels predominantly select for color and low-level texture; V2 is dominated by texture; V4 begins producing parts and object fragments (wheel, head, ear); IT is dominated by whole objects.

\begin{figure}[t]
  \centering
  \includegraphics[width=\linewidth]{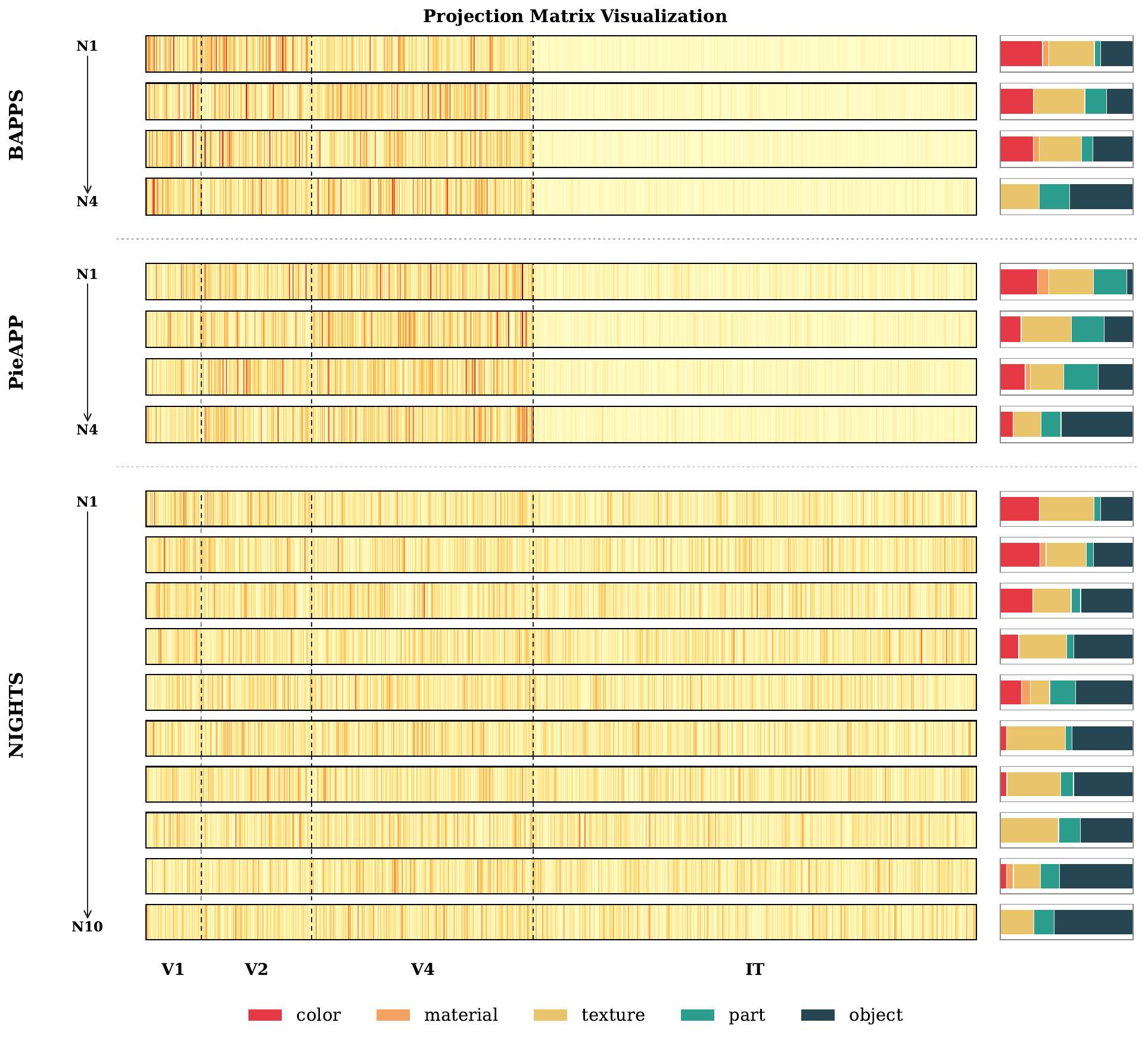}
  \caption{
Projection structure of the learned perceptual dimensions.
We visualize the projection matrix for BAPPS ($N=4$), PieAPP ($N=4$), and NIGHTS ($N=10$).
The heatmaps show row-normalized projection magnitudes $|W_{k,j}|$, where each row corresponds to one learned perceptual dimension and each column to one of the $960$ CORnet-S-Multi channels.
Dashed vertical lines mark the boundaries between V1, V2, V4, and IT channels.
For each row, the bar on the right summarizes the BRODEN category composition of the top-$20$ weighted channels.
The five segments, color-coded according to the bottom legend, correspond to color, material, texture, part, and object, and sum to one.
Together, the heatmaps indicate where each dimension reads from in the visual hierarchy, while the side bars summarize what visual content those channels encode.
}
  \label{fig:appendix-weights}
\end{figure}

\paragraph{Perceptual Dimension Interpretation.}
The learned projection matrix $\mathbf{W} \in \mathbb{R}^{N \times D}$ identifies which of these feature channels each perceptual dimension reads from.
\Fig{fig:appendix-weights} shows the projection matrix for the three datasets, BAPPS at $N=4$, PieAPP at $N=4$, and NIGHTS at $N=10$, using CORnet-S Multi.
The project matrices are row normalized and visualized as heatmaps.
BAPPS and PieAPP models do not rely on area IT, whose contribution becomes more significant for NIGHTS.
The level of the ventral stream recruited by the perceptual space therefore tracks the semantic level of the dataset.

Each row of the project matrix represents a perceptual dimension.
We pick the top-20 feature channels of each dimension, finds the top-3 concept associated with each channel, and aggregate across all the channels to plot a concept distribution of each perceptual dimension.
The distribution plots are visualized alongside the projection matrices.

From the distribution plots, we can see that at higher $N$ the concept associated with of each dimension becomes more diverse and mixed:
BAPPS and PieAPP concentrate weight on V1, V2, and V4, and their dimensions remain dominated by color and texture;
NIGHTS develops a substantial dependence on IT, and its dimensions reach part- and object-level semantics.
Higher dimensionality is therefore not redundant repetition but a finer partition of the same hierarchy into specialized subspaces.

\paragraph{Implementation Details.}
For each CORnet-S channel $j$, we binarize its activation at the top-$0.5\%$ threshold (estimated on a $5{,}000$-image subsample, matching \citet{bau2017network}) and upsample to BRODEN's $112\times 112$ resolution to obtain mask $M_j(x)$.
For each concept $c$ with ground-truth mask $L_c(x)$, we compute $\text{IoU}(j, c) = \sum_x |M_j \cap L_c| / \sum_x |M_j \cup L_c|$ and divide by a closed-form random-IoU baseline; this ratio puts color and texture on the same scale and is what we call a channel's selectivity for the concept.
A channel is considered aligned with a concept when its selectivity passes $5\times$.
The dissection produces a top-3 concept list per channel ranked by raw IoU; we take the entry with the largest selectivity as the channel's representative concept, since selectivity reflects discriminative power better than raw overlap.
For each CORnet-S-Multi checkpoint with projection $\mathbf{W} \in \R^{n \times 960}$, we sort channels by $|W_{k,j}|$, take the top $20$, and aggregate their representative concepts into the category-mix bars in \Fig{fig:appendix-weights}.

\subsection{Dimensions are Independent}
\label{app:perceptual-corr}

\begin{figure}[t]
  \centering
  \includegraphics[width=\linewidth]{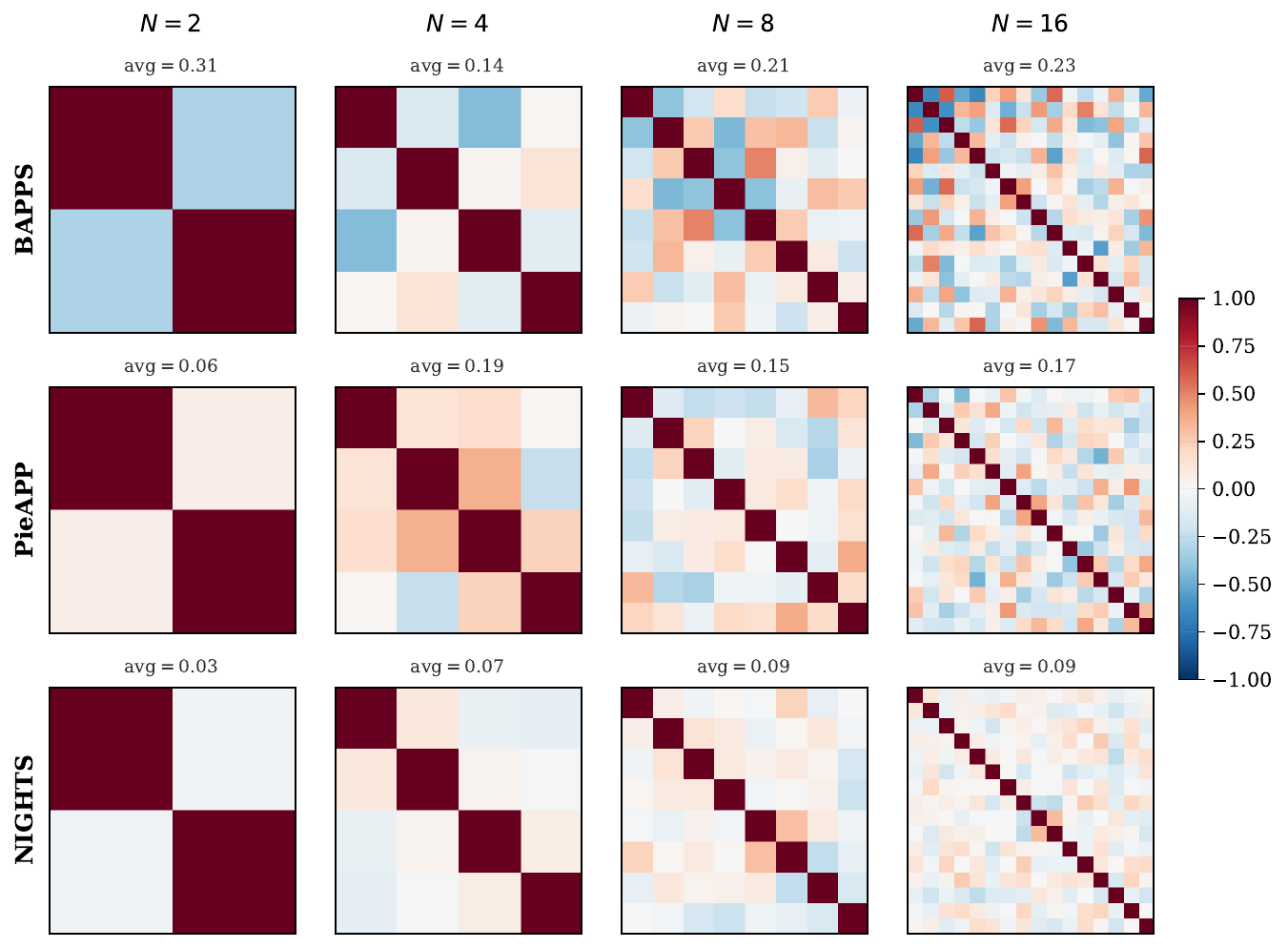}
  \caption{Pairwise Pearson correlation between learned dimensions on the validation set.
  Rows are BAPPS, PieAPP and NIGHTS.
  Columns are $N=2,4,8,16$.
  The number above each cell is the average
  off-diagonal coefficient.
  }
  \label{fig:appendix-correlation}
\end{figure}

Finally, we empirically demonstrate that the learned perceptual dimensions are mutually orthogonal within the perceptual space, a property that is not guaranteed a priori.
We use a method similar to that in \citet{zbontar2021barlow}:
each image in the dataset, assuming that are $I$ of them, is mapped to a N-dimensional vector in the perceptual space $\mathbb{R}^N$, form a $N \times I$ matrix;
in other words, each dimension is represented by a $V$-dimensional vector.
We calculate the pairwise Pearson correlation between every dimension pair, and plot the results in \Fig{fig:appendix-correlation}.
Across all $N$s and all datasets, the dimensions are largely independent of each other, with an average off-diagonal correlation coefficient smaller than 0.25.

\section{Results on Different Feature Encoders and Projection Choices}
\label{app:full-results}
\label{app:indomain-full}

\Tbl{tab:app-indomain} reports the prediction performance under different feature encoders.
We also evaluate two dimensionality methods: learned projection (L), which is used in the main text, and PCA (P).
Shaded rows indicate the encoders used in the main text and all cells show peak performance and the corresponding saturating $\nstar$.

\begin{table}[h]
\centering
\caption{Full in-domain results across three benchmarks under two projection configurations (L = Learned, P = PCA).
BAPPS: agreement/$\nstar$ and KL/$\nstar$ ($0$ = oracle).
NIGHTS: accuracy/$\nstar$. PieAPP: agreement/$\nstar$.
Theoretical ceilings reported in main-text \Tbl{tab:baseline}.}
\label{tab:app-indomain}
\scriptsize
\setlength{\tabcolsep}{2pt}
\begin{tabular}{l cc cc cc cc}
\toprule
& \multicolumn{2}{c}{BAPPS Agr / $\nstar$} & \multicolumn{2}{c}{BAPPS KL / $\nstar$} & \multicolumn{2}{c}{NIGHTS Acc / $\nstar$} & \multicolumn{2}{c}{PieAPP Agr / $\nstar$} \\
\cmidrule(lr){2-3} \cmidrule(lr){4-5} \cmidrule(lr){6-7} \cmidrule(lr){8-9}
Encoder & L & P & L & P & L & P & L & P \\
\midrule
VGG-conv3 ($4{\times}4$)
  & .688/7 & .667/10
  & .174/12 & .186/10
  & .797/32 & .771/14
  & .711/9 & .674/4 \\
VGG-conv3
  & .687/4 & .666/9
  & .170/5 & .185/9
  & .802/16 & .786/16
  & .709/6 & .677/3 \\
VGG-conv5 ($2{\times}2$)
  & .666/7 & .649/8
  & .189/6 & .203/5
  & .801/48 & .769/28
  & .689/20 & .678/12 \\
VGG-conv5
  & .670/7 & .648/7
  & .184/7 & .201/7
  & .831/96 & .794/48
  & .687/8 & .675/10 \\
VGG-multi
  & .692/4 & .671/9
  & .164/7 & .181/12
  & .848/24 & .811/24
  & .709/6 & .677/3 \\
\addlinespace
\rowcolor[gray]{0.95}
CORnet-S V1
  & .685/4 & .669/14
  & .175/6 & .189/8
  & .747/8 & .741/7
  & .701/6 & .689/14 \\
\rowcolor[gray]{0.95}
CORnet-S V2
  & .689/3 & .672/12
  & .166/5 & .181/12
  & .793/14 & .780/12
  & .712/5 & .682/16 \\
\rowcolor[gray]{0.95}
CORnet-S V4
  & .688/4 & .662/8
  & .162/8 & .189/8
  & .855/48 & .823/20
  & .714/8 & .682/3 \\
\rowcolor[gray]{0.95}
CORnet-S IT
  & .677/6 & .662/10
  & .180/6 & .191/9
  & .853/96 & .840/48
  & .702/14 & .684/24 \\
\rowcolor[gray]{0.95}
CORnet-S Multi
  & .688/3 & .664/12
  & .167/4 & .188/10
  & .859/96 & .838/48
  & .710/9 & .682/16 \\
\bottomrule
\end{tabular}
\end{table}

The conclusion from \Sect{sec:tasks}---that the perceptual space underlying low-level quality judgments is much lower-dimensional than the perceptual space underlying high-level judgments---holds across both projection configurations and across every encoder we evaluate.
Under the learned projection, $\nstar$ on BAPPS sits in the $3$--$7$ range under the agreement metric, and similarly low under KL divergence, so the saturation diagnosis does not depend on a particular choice of metric.
PCA's performance is universally lower than that of the learned projection, showing the importance of the latter.

\section{Results on Another High-Level Dataset}
\label{app:deep-ranking}

NIGHTS is the only high-level dataset in the main text.
To test whether its higher saturating dimensionality is specific to NIGHTS, we additionally fit our observer model on Deep-Ranking~\citep{wang2014learning}, a high-level 2AFC triplet dataset built from web image search.
We use the same feature encoders and protocol as for NIGHTS.

\begin{table}[h]
\centering
\caption{Accuracy\,/\,$\nstar$ on the two high-level datasets. NIGHTS results are the same as in \Tbl{tab:main}.}
\label{tab:app-deep-ranking}
\scriptsize
\setlength{\tabcolsep}{4pt}
\begin{tabular}{@{}l cc@{}}
\toprule
Feature Encoder & NIGHTS & Deep-Ranking \\
\midrule
CORnet-S V1    & .747\,/\,8  & .667\,/\,8 \\
CORnet-S V2    & .793\,/\,14 & .713\,/\,8 \\
CORnet-S V4    & .855\,/\,48 & .787\,/\,24 \\
CORnet-S IT    & .853\,/\,96 & .821\,/\,96 \\
CORnet-S Multi & .859\,/\,96 & .814\,/\,96 \\
\bottomrule
\end{tabular}
\end{table}

\Tbl{tab:app-deep-ranking} shows that Deep-Ranking follows the same pattern as NIGHTS.
Performance improves from V1 to higher visual areas, and with IT and CORnet-S Multi the saturating dimensionality reaches $96$ on both datasets, an order of magnitude higher than on BAPPS and PieAPP (\Tbl{tab:main}).
Thus, the higher dimensionality reflects the high-level nature of the task rather than the distribution of a particular dataset.

\section{Results on Cross-Dataset Prediction}
\label{app:cross-results}

\begin{figure}[h]
\centering
\includegraphics[width=0.92\linewidth]{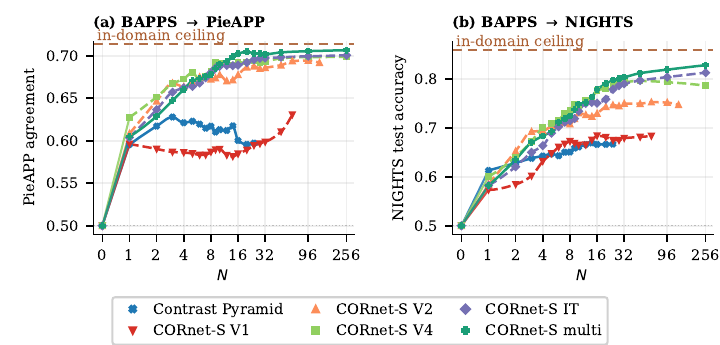}
\caption{Cross-dataset transfer of a BAPPS-trained observer across six feature encoders.
(a): BAPPS\,$\rightarrow$\,PieAPP using per-annotator agreement on PieAPP as the accuracy metric.
(b): BAPPS\,$\rightarrow$\,NIGHTS using 1-BER on the NIGHTS test set.
The brown dashed line in each panel is the best in-domain ceiling on the test-side dataset.
}
\label{fig:app-cross-domain}
\end{figure}

\Fig{fig:app-cross-domain} quantifies the cross-dataset claim of \Sect{sec:results}.
For each encoder we train the observer on BAPPS, freeze it, and evaluate it on a different dataset without any further fitting.
The brown dashed line in each panel is the best in-domain peak on the target dataset, i.e., that of a model trained directly on PieAPP or NIGHTS.
Transferring to PieAPP (\Fig{fig:app-cross-domain}(a)), another low-level 2AFC dataset, costs the high-level CORnet-S encoder (V4, IT, multi-layer) almost nothing: their curves almost meet the in-domain ceiling.
Transferring to NIGHTS (\Fig{fig:app-cross-domain}(b)), in contrast, leaves every encoder several points below the in-domain ceiling, and the residual gap does not close as $N$ grows or as we move up the visual hierarchy.

\Tbl{tab:app-cross} reports the corresponding peaks and saturation points across all encoders.
The cross-domain pattern mirrors the in-domain pattern, with $\nstar$ on the target dataset often higher than its in-domain counterpart.

\begin{table}[h]
\centering
\caption{Full cross-domain results under the default Learned projection with Cov$\leftarrow\bm{\mu}(x)$. B$\to$N: trained on BAPPS, tested on NIGHTS (accuracy/$\nstar$). N$\to$B: trained on NIGHTS, tested on BAPPS (agreement/$\nstar$). B$\to$P: trained on BAPPS, tested on PieAPP (agreement/$\nstar$). Theoretical ceilings reported in main-text \Tbl{tab:baseline}. Shaded rows are the encoders used in the main text.}
\label{tab:app-cross}
\scriptsize
\setlength{\tabcolsep}{4pt}
\begin{tabular}{l c c c}
\toprule
Encoder & B$\to$N: NIGHTS Acc / $\nstar$ & N$\to$B: BAPPS Agr / $\nstar$ & B$\to$P: PieAPP Agr / $\nstar$ \\
\midrule
VGG-conv3 ($4{\times}4$)        & .735/32  & .669/10 & .702/9 \\
VGG-conv3                       & .770/32  & .653/28 & .697/14  \\
VGG-conv5 ($2{\times}2$)        & .714/24  & .638/20 & .685/20 \\
VGG-conv5                       & .752/48  & .626/24  & .684/12 \\
VGG-multi                       & .811/48  & .670/12  & .688/5  \\
\addlinespace
\rowcolor[gray]{0.95}
CORnet-S V1                       & .688/16  & .639/8  & .631/64  \\
\rowcolor[gray]{0.95}
CORnet-S V2                       & .757/20  & .659/10  & .694/20 \\
\rowcolor[gray]{0.95}
CORnet-S V4                       & .796/20  & .663/28 & .701/10 \\
\rowcolor[gray]{0.95}
CORnet-S IT                       & .813/96 & .666/48 & .701/20 \\
\rowcolor[gray]{0.95}
CORnet-S Multi                    & .825/48  & .670/24 & .707/14 \\
\bottomrule
\end{tabular}
\end{table}

\section{Extension to Rating-Based IQA Datasets}
\label{app:friqa}

Our observer model can be readily extended to rating-based IQA datasets, where each image is associated with a perceptual score.
In our setting, the \textit{expected} Minkowski distance between an image's distribution and that of the reference image in the learned perceptual space can be treated as the perceptual score of the image.

We train the observer model (CORnet-S-Multi encoder, $N{=}24$) end-to-end on the full KADID-10k MOS labels with an $L_1 + (1{-}\rho)$ loss, where $\rho$ is the Pearson correlation between predicted and ground-truth quality.
A two-parameter sigmoid score head $q = \sigma\!\left(a\,\log(1{+}d) + b\right)$ with $a, b$ learnable maps the weighted Minkowski distance $d$ of \Equ{eq:distance} to a scalar predicted MOS-badness $q \in [0, 1]$; this affine head reshapes the heavy tail of $d$ without breaking monotonicity.
Per-pair distances are computed by Monte Carlo with $K{=}512$ samples;
the Spearman rank-order correlation coefficients (SRCC), Pearson linear correlation coefficients (PLCC), and Kendall rank correlation coefficients (KRCC) are reported as absolute values per the convention, with PLCC computed after the 4-parameter logistic remapping recommended by the Video Quality Experts Group (VQEG)~\citep{rohaly2000video} to align predicted scores with the subjective MOS scale before linear comparison.

We evaluate the resulting model's zero-shot on three standard rating-based IQA benchmarks: TID2013, CSIQ, and LIVE Release~2, all of which contain traditional algorithmic distortions (JPEG, JPEG2000, blur, noise, colour/contrast shifts) similar in kind to KADID-10k.
KADID-10k itself is excluded from the comparison because our MOS-fit model is trained on its full set; for the same reason, TOPIQ-FR's released checkpoint, which was trained by \citet{chen2024topiq} on the full KADID-10k and is the public benchmark used here, would also produce training-set numbers on KADID-10k and is omitted on that dataset.
Baseline numbers for PSNR, SSIM, LPIPS-VGG, DISTS, DreamSim, PieAPP, CVVDP, MILO, and TOPIQ-FR are computed with the same evaluation protocol.

\begin{table}[h]
\centering
\small
\caption{The observer model extends to rating-based IQA datasets with absolute (D)MOS labels. Each cell reports SRCC\,/\,PLCC\,/\,KRCC; higher is better for all three. \textbf{Bold} = best, \uline{underline} = second \& third best, ranked independently per (dataset, metric) column.}
\label{tab:app-friqa}
\begin{tabular}{@{}lccc@{}}
\toprule
Method & TID2013 & CSIQ & LIVE \\
\midrule
CORnet-S Multi                                 & \uline{0.860} / \uline{0.870} / 0.666 & \uline{0.946} / \uline{0.948} / \uline{0.796} & 0.937 / \uline{0.927} / 0.781 \\
\midrule
MILO                                                & \uline{0.876} / \uline{0.876} / \uline{0.692} & \uline{0.959} / \uline{0.942} / \uline{0.817} & \uline{0.953} / 0.913 / \textbf{0.817} \\
TOPIQ-FR                                            & \textbf{0.917} / \textbf{0.921} / \textbf{0.744} & \textbf{0.967} / \textbf{0.964} / \textbf{0.838} & 0.898 / 0.901 / 0.717 \\
CVVDP                                               & 0.853 / 0.864 / \uline{0.672} & 0.896 / 0.883 / 0.728 & 0.917 / 0.902 / 0.766 \\
DreamSim                                            & 0.809 / 0.830 / 0.612 & 0.932 / 0.941 / 0.768 & 0.899 / 0.902 / 0.729 \\
PieAPP                                              & 0.850 / 0.846 / 0.652 & 0.845 / 0.829 / 0.666 & 0.925 / 0.915 / 0.765 \\
DISTS                                               & 0.708 / 0.755 / 0.521 & 0.930 / 0.938 / 0.764 & \textbf{0.955} / \textbf{0.954} / \uline{0.814} \\
LPIPS-VGG                                           & 0.670 / 0.749 / 0.497 & 0.883 / 0.905 / 0.697 & \uline{0.943} / \uline{0.945} / \uline{0.791} \\
SSIM                                                & 0.710 / 0.702 / 0.511 & 0.813 / 0.804 / 0.613 & 0.927 / 0.910 / 0.762 \\
PSNR                                                & 0.687 / 0.677 / 0.496 & 0.809 / 0.816 / 0.599 & 0.925 / 0.924 / 0.756 \\
\bottomrule
\end{tabular}
\end{table}

\paragraph{Results.}
\Tbl{tab:app-friqa} shows the results.
Across all three datasets and all three metrics, our method often ranks among the top three compared the baselines.
This shows that our observer model can well account for human data even on rating-based IQA datasets.

We summarize the results on both triplet 2AFC datasets (\Tbl{tab:baseline}) and the rating-based IQA datasets (\Tbl{tab:app-friqa}) into a single table \Tbl{tab:app-rank}, where we show the average ranking of a method across all six datasets.
Our method maintains strong performance across both settings.
It achieves the best average rank of $2.83$ and ranks within the top three on five of the six datasets.

\begin{table}[h]
\centering
\small
\caption{Per-dataset rank of each method across the six datasets covered by Tables~\ref{tab:baseline} and~\ref{tab:app-friqa}, restricted to the nine baselines that appear in both. Score per dataset: BAPPS \& PieAPP use per-annotator agreement; NIGHTS uses accuracy; TID2013 \& CSIQ \& LIVE use SRCC. Lower rank is better; the right-most column is the mean across the six datasets.
}
\label{tab:app-rank}
\begin{tabular}{@{}lcccccc|c@{}}
\toprule
Method & BAPPS & NIGHTS & PieAPP & TID2013 & CSIQ & LIVE & avg.\ rank \\
\midrule
CORnet-S Multi & 1 & 3 & 3 & 3 & 3 & 4 & \textbf{2.83} \\
\midrule
DreamSim    & 2 & 1 & 1 & 6 & 4 & 9 & 3.83 \\
DISTS       & 3 & 2 & 4 & 8 & 5 & 1 & 3.83 \\
TOPIQ-FR    & 4 & 6 & 5 & 1 & 1 & 10 & 4.50 \\
MILO        & 8 & 9 & 7 & 2 & 2 & 2 & 5.00 \\
LPIPS       & 5 & 4 & 6 & 10 & 7 & 3 & 5.83 \\
CVVDP       & 6 & 5 & 8 & 4 & 6 & 8 & 6.17 \\
PieAPP      & 10 & 7 & 2 & 5 & 8 & 6 & 6.33 \\
SSIM        & 7 & 8 & 9 & 7 & 9 & 5 & 7.50 \\
PSNR        & 9 & 10 & 10 & 9 & 10 & 7 & 9.17 \\
\bottomrule
\end{tabular}
\end{table}

\begin{figure}[h]
\centering
\includegraphics[width=\linewidth]{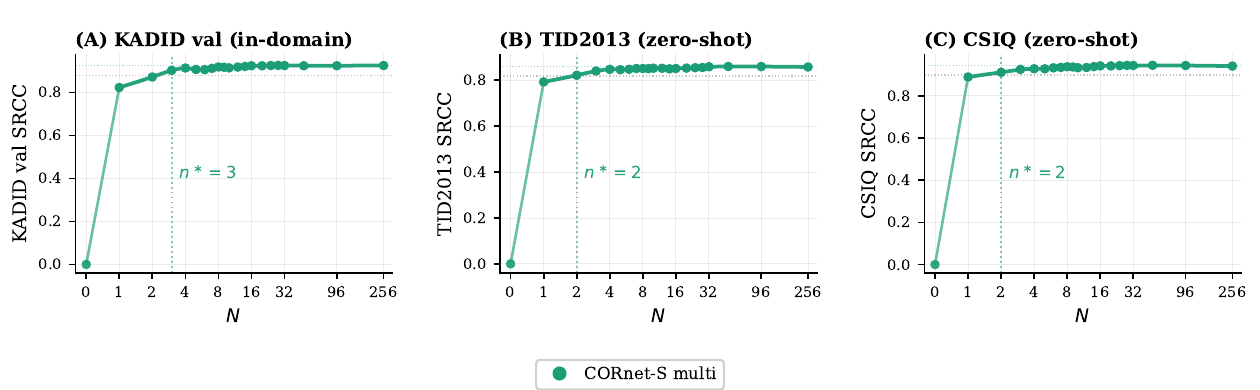}
\caption{Saturation of the KADID fitted CORnet-Multi observer model ($N{=}24$ in \Tbl{tab:app-friqa}) along the sweep schedule. \textbf{(A)} KADID-10k val. \textbf{(B)} TID2013 (zero-shot). \textbf{(C)} CSIQ (zero-shot). The vertical band marks the 95\%-of-peak saturation point $n^{\star}$, computed against an SRCC chance baseline of $0$.}
\label{fig:app-mosprobe-saturation}
\end{figure}

\paragraph{Saturation Dimensionality Persists.}
\Fig{fig:app-mosprobe-saturation} plots SRCC at every swept $N$ for the CORnet-S Multi observer reported in \Tbl{tab:app-friqa}, on KADID-10k val (20\% by-reference held-out) and on the two zero-shot test sets, TID2013 and CSIQ.
The low-dimensional saturation pattern observed in the main text holds.
The curve of KADID-10k saturates at $n^{\star} = 3$,
while TID2013 and CSIQ saturate at $n^{\star} = 2$; SRCC then fluctuates by less than $\pm 0.005$ across $n \in \{24, 32, 48, 96, 256\}$.
Because TID2013, CSIQ, and KADID-10k all consist of traditional low-level distortions (JPEG, JPEG2000, blur, noise, color/contrast shifts), this results here corroborate the findings on the triplet 2AFC tasks (\Sect{sec:tasks}).

\section{Comparison with Bradley--Terry Model}
\label{app:bt}

The BT model~\cite{bradley1952rank} is commonly used to relate quality scores of images/objects to choice probabilities in pairwise comparisons.
Given two images, each with a quality score $s_i$ and $s_j$, the probability of choosing $i$ over $j$ (under a reference $r$) is modeled to be: $\Pr(i|r, i, j) = \frac{1}{1+e^{s_j-s_i}}$.

One can derive this model by assuming that the quality score of each image is a random variable $S_i$, which has mean $s_i$ and follows a certain distribution: $S_i \sim \mathcal{D}$.
In a pairwise comparison, the probability that image $i$ is picked over image $j$ (with respect to the reference $r$) is then the expectation that $S_i - S_j > 0$.
That is, $\Pr(i \mid r, i, j) \;=\; \mathbb{E}\!\big[\, \mathbf{1}\{S_j < S_i\} \,\big]$.
One can show that when the distribution $\mathcal{D}$ is a Gumbel distribution with a constant standard deviation, this formulation is exactly the BT model~\cite{tsukida2011analyze};
when $\mathcal{D}$ is a Gaussian distribution with, again, a constant standard deviation, this formulation is equivalent to the Thurstone model Case V~\cite{thurstone1927law}.

Note the similarity between this formulation and \Equ{eq:choice}.
The near-identical forms are not coincidental; it is because our observer model, while modeling the distribution of high-dimensional, internal responses of an image, can also be equivalently interpreted as modeling the one-dimensional, perceptual-score distribution of an image.
The score can be given by the Minkowski distance between the image and the reference in the perceptual space (\Equ{eq:distance}).
The distribution of this perceptual score, however, is neither Gumbel nor Gaussian, even if the perceptual space itself is Gumbel or Gaussian, as it involves calculating the Minkowski distance.
Alternatively, one can, like \citet{schmidt2025material}, map each image into an embedding and use the cosine similarity between the embedding of an image and that of a reference as the score, and simply assume that such a score is a drawn from a Gumbel random variable.

Our implementation of the BT model uses the latter approach.
Specifically, we strip the multi-dimensional, image-dependent noise from the observer model while keeping the $N$-dimensional latent embedding $\bm{\mu}(x)$ and the learned projection $\mathbf{W}$ intact, so that the only change from the full observer model is the noise parameterization.
We then replace the weighted Minkowski distance of \Equ{eq:distance} with the cosine similarity between each candidate's latent and the reference's embedding.
Writing $s_i \;=\; \tau\,\cos\!\big(\bm{\mu}(r), \bm{\mu}(x_i)\big)$ for the score of candidate $i$ scaled by a learnable temperature $\tau$,
the probability of choosing $x_0$ is simply $\frac{1}{1+e^{s_1-s_0}}$, as given by the BT model.
The underlying assumption is that $s_0$ and $s_1$ both follow a Gumbel distribution that differ only in the mean.

\Tbl{tab:app-bt-indomain} and~\ref{tab:app-bt-cross} compare the two methods (our full observer model and the BT model) across 10 feature encoders, both in-domain and cross-domain.
Experimentally, we find that the BT model and our method give similar predictive power.
By using physiologically grounded feature encoders, however, our method permits examining (the dimensions of) the internal perceptual space, which the original BT model is incapable of\footnote{We make a distinction between the original BT model~\cite{bradley1952rank}, where each image/object is directly associated with a score that is directly estimated through maximum likelihood, and the BT model augmented by a multi-dimensional observer model, like the one we present here, where the score is derived from a high-dimensional perceptual space. In the latter case, one could in principle still study the structure of the perceptual space (e.g., the saturating dimensionality).}, and can be thought of as giving a low-level interpretation of the phenomenological BT model.

\begin{table}[h]
\centering
\caption{BT vs.\ full observer model, in-domain. $\Delta$ = BT $-$ Obs. BAPPS: per-annotator agreement. NIGHTS: 2AFC accuracy.}
\label{tab:app-bt-indomain}
\scriptsize
\setlength{\tabcolsep}{2pt}
\begin{tabular}{lr ccccc ccccc}
\toprule
& & \multicolumn{5}{c}{BAPPS (Agr)} & \multicolumn{5}{c}{NIGHTS (Acc)} \\
\cmidrule(lr){3-7} \cmidrule(lr){8-12}
Backbone & $D$ & BT & Obs & $\Delta$ & BT $\nstar$ & Obs $\nstar$ & BT & Obs & $\Delta$ & BT $\nstar$ & Obs $\nstar$ \\
\midrule
CORnet-S V1        & 64   & .686 & .685 & $+.001$ & 4 & 4  & .764 & .747 & $+.017$ & 12  & 8   \\
CORnet-S V2        & 128  & .691 & .689 & $+.002$ & 7 & 3  & .783 & .793 & $-.010$ & 12  & 14  \\
CORnet-S V4        & 256  & .687 & .688 & $-.001$ & 6 & 4  & .831 & .855 & $-.024$ & 48  & 48  \\
CORnet-S IT        & 512  & .677 & .677 & $+.000$ & 8 & 6  & .862 & .853 & $+.009$ & 96  & 96  \\
CORnet-S Multi     & 960  & .685 & .688 & $-.003$ & 6 & 3  & .859 & .859 & $+.000$ & 256  & 96  \\
VGG-conv3    & 256  & .688 & .687 & $+.001$ & 5 & 4  & .797 & .802 & $-.005$ & 20  & 16  \\
VGG-conv5    & 512  & .671 & .670 & $+.001$ & 6 & 7  & .837 & .831 & $+.006$ & 48 & 96  \\
VGG-conv3 ($4{\times}4$)  & 4096 & .679 & .688 & $-.009$ & 5 & 7  & .777 & .797 & $-.020$ & 9  & 32  \\
VGG-conv5 ($2{\times}2$)  & 2048 & .664 & .666 & $-.002$ & 8 & 7  & .826 & .801 & $+.025$ & 256  & 48  \\
VGG-multi  & 1472 & .686 & .692 & $-.006$ & 5 & 4  & .827 & .848 & $-.021$ & 24  & 24  \\
\midrule
\textbf{Average} &      &      &      & $\mathbf{-.002}$ & & &      &      & $\mathbf{-.002}$ & & \\
\bottomrule
\end{tabular}
\end{table}

\begin{table}[h]
\centering
\caption{BT vs.\ full observer model, cross-domain. $\Delta$ = BT $-$ Obs. B$\to$N: 2AFC accuracy. N$\to$B: per-annotator agreement.}
\label{tab:app-bt-cross}
\scriptsize
\setlength{\tabcolsep}{2.5pt}
\begin{tabular}{lr ccc ccc}
\toprule
& & \multicolumn{3}{c}{B$\to$N (Acc)} & \multicolumn{3}{c}{N$\to$B (Agr)} \\
\cmidrule(lr){3-5} \cmidrule(lr){6-8}
Backbone & $D$ & BT & Obs & $\Delta$ & BT & Obs & $\Delta$ \\
\midrule
CORnet-S V1        & 64   & .712 & .688 & $+.024$ & .661 & .639 & $+.022$ \\
CORnet-S V2        & 128  & .763 & .757 & $+.006$ & .656 & .659 & $-.003$ \\
CORnet-S V4        & 256  & .823 & .796 & $+.027$ & .661 & .663 & $-.002$ \\
CORnet-S IT        & 512  & .847 & .813 & $+.034$ & .664 & .666 & $-.002$ \\
CORnet-S Multi     & 960  & .837 & .825 & $+.012$ & .665 & .670 & $-.005$ \\
VGG-conv3    & 256  & .785 & .770 & $+.015$ & .646 & .653 & $-.007$ \\
VGG-conv5    & 512  & .802 & .752 & $+.050$ & .654 & .626 & $+.028$ \\
VGG-conv3 ($4{\times}4$)  & 4096 & .725 & .735 & $-.010$ & .647 & .669 & $-.022$ \\
VGG-conv5 ($2{\times}2$)  & 2048 & .772 & .714 & $+.058$ & .653 & .638 & $+.015$ \\
VGG-multi  & 1472 & .815 & .811 & $+.004$ & .654 & .670 & $-.016$ \\
\midrule
\textbf{Average} &      &      &      & $\mathbf{+.022}$ &      &      & $\mathbf{+.001}$ \\
\bottomrule
\end{tabular}
\end{table}

\section{MAP Using Wishart Prior on the Covariance}
\label{app:wishart}

The main text fits the observer model by maximum likelihood.
One can also apply a prior on the covariance matrix field.
In particular, the Wishart distribution is widely used as a probabilistic model for covariance ~\cite{hong2025comprehensive, nejatbakhsh2023estimating}.
We impose a weak prior that each covariance matrix is a sample from a Wishart distribution, and integrate the prior with the likelihood function using maximum a posteriori (MAP) estimation.

\subsection{MAP with Wishart Distribution Prior}

The Wishart distribution $\mathcal{W}_n(\mathbf{V}, \nu)$ is a distribution over $n \times n$ positive semi-definite matrices, parameterized by a scale matrix $\mathbf{V} \in \mathbb{S}^n_+$ and degrees of freedom $\nu \geq n$.
Its probability density function is:
\begin{equation}
  \label{eq:wishart-pdf}
  p(\bm{\Sigma}) = \frac{|\bm{\Sigma}|^{(\nu - n - 1)/2} \exp\!\left(-\tfrac{1}{2} \mathrm{tr}(\mathbf{V}^{-1} \bm{\Sigma})\right)}{2^{\nu n/2}\, |\mathbf{V}|^{\nu/2}\, \Gamma_n(\nu/2)},
\end{equation}
where $\Gamma_n(\cdot)$ is the multivariate gamma function $\Gamma_n(a) = \pi^{n(n-1)/4} \prod_{j=1}^{n} \Gamma(a - (j{-}1)/2)$.

We fix $\nu = n + 2$ (the minimal value above $n + 1$ that gives a positive log-det coefficient).
We choose $\mathbf{V} = \tfrac{1}{\nu}\mathbf{I}$, which yields an isotropic prior with mean $\mathbb{E}[\bm{\Sigma}] = \mathbf{I}$.
Both $\nu$ and $\bm{V}$ can in theory be learned, which we leave to future work.

With the prior, the optimization problem becomes MAP:
\begin{equation}
  \label{eq:map}
  \hat\Theta \;=\; \arg\max_{\Theta}(\log p(\mathcal{D} \mid \Theta) \;+\; \lambda\,\log p(\Theta)),
\end{equation}
with $\lambda \geq 0$ controlling how strongly the prior is enforced relative to the likelihood (if the Bayes rule is to be strictly followed, $\lambda = 1$).
The first term is the log likelihood function in \Equ{eq:loglik}.
The second term is the prior defined over the free parameters $\Theta$, and can be trivially derived by plugging \Equ{eq:chol} and \Equ{eq:mlp} into \Equ{eq:wishart-pdf};
we omit the exact express for simplicity.

\subsection{Results}

We compare plain maximum likelihood ($\lambda = 0$, baseline) against MAP with $\lambda = 0.05$ and $\lambda = 1.0$ (strict MAP) across 10 feature encoders.
\Tbl{tab:app-wishart-indomain} shows the results on BAPPS and NIGHTS (in-domain tests), and~\Tbl{tab:app-wishart-cross} shows the results when a BAPPS-trained model is applied to NIGHTS and vice versa (cross-domain tests; similar to those in \Sect{app:cross-results}).
$\Delta$ shows the prediction performance gain from the using the prior.

There is a small net negative for in-domain tests ($-0.3\%$ BAPPS agreement, $\Delta = -0.1\%$ on NIGHTS for $\lambda = 0.05$ and $-0.5\%$ for $\lambda = 1.0$), as expected from a prior that trades training-set fit for less extreme covariances;
the dimensionality finding itself is unchanged.
The effects on cross-domain tests is roughly neutral on average.
Overall, the Wishart prior delivers no consistent gain in either test.
This is likely due to the fact that the baseline MLE method is jointly learning both the mean field and the covariance matrix, and any potential effects of the priors on the covariance matrix field could be learned through the basic MLE itself already.

\begin{table}[h]
\centering
\caption{Wishart MAP vs.\ MLE in-domain. $\Delta$ = MAP $-$ MLE. BAPPS: per-annotator agreement. NIGHTS: 2AFC accuracy.}
\label{tab:app-wishart-indomain}
\scriptsize
\setlength{\tabcolsep}{2pt}
\begin{tabular}{lr ccccc ccccc}
\toprule
& & \multicolumn{5}{c}{BAPPS (Agr)} & \multicolumn{5}{c}{NIGHTS (Acc)} \\
\cmidrule(lr){3-7} \cmidrule(lr){8-12}
Backbone & $D$ & MLE & $\lambda{=}.05$ & $\Delta$ & $\lambda{=}1$ & $\Delta$ & MLE & $\lambda{=}.05$ & $\Delta$ & $\lambda{=}1$ & $\Delta$ \\
\midrule
CORnet-S V1        & 64   & .685 & .668 & $-.017$ & .670 & $-.015$ & .747 & .737 & $-.010$ & .732 & $-.015$ \\
CORnet-S V2        & 128  & .689 & .683 & $-.006$ & .683 & $-.006$ & .793 & .788 & $-.005$ & .782 & $-.011$ \\
CORnet-S V4        & 256  & .688 & .686 & $-.002$ & .686 & $-.002$ & .855 & .839 & $-.016$ & .835 & $-.020$ \\
CORnet-S IT        & 512  & .677 & .676 & $-.001$ & .676 & $-.001$ & .853 & .855 & $+.002$ & .845 & $-.008$ \\
CORnet-S Multi     & 960  & .688 & .688 & $+.000$ & .688 & $+.000$ & .859 & .868 & $+.009$ & .861 & $+.002$ \\
VGG-conv3    & 256  & .687 & .686 & $-.001$ & .687 & $+.000$ & .802 & .811 & $+.009$ & .807 & $+.005$ \\
VGG-conv5    & 512  & .670 & .667 & $-.003$ & .669 & $-.001$ & .831 & .832 & $+.001$ & .827 & $-.004$ \\
VGG-conv3 ($4{\times}4$)  & 4096 & .688 & .686 & $-.002$ & .685 & $-.003$ & .797 & .798 & $+.001$ & .798 & $+.001$ \\
VGG-conv5 ($2{\times}2$)  & 2048 & .666 & .665 & $-.001$ & .666 & $+.000$ & .801 & .800 & $-.001$ & .808 & $+.007$ \\
VGG-multi  & 1472 & .692 & .692 & $+.000$ & .692 & $+.000$ & .848 & .846 & $-.002$ & .845 & $-.003$ \\
\midrule
\textbf{Average} &      &      &      & $\mathbf{-.003}$ &      & $\mathbf{-.003}$ &      &      & $\mathbf{-.001}$ &      & $\mathbf{-.005}$ \\
\bottomrule
\end{tabular}
\end{table}

\begin{table}[h]
\centering
\caption{Wishart MAP vs.\ MLE cross-domain. $\Delta$ = MAP $-$ MLE. B$\to$N: 2AFC accuracy. N$\to$B: per-annotator agreement.}
\label{tab:app-wishart-cross}
\scriptsize
\setlength{\tabcolsep}{2pt}
\begin{tabular}{lr ccccc ccccc}
\toprule
& & \multicolumn{5}{c}{B$\to$N (Acc)} & \multicolumn{5}{c}{N$\to$B (Agr)} \\
\cmidrule(lr){3-7} \cmidrule(lr){8-12}
Backbone & $D$ & MLE & $\lambda{=}.05$ & $\Delta$ & $\lambda{=}1$ & $\Delta$ & MLE & $\lambda{=}.05$ & $\Delta$ & $\lambda{=}1$ & $\Delta$ \\
\midrule
CORnet-S V1        & 64   & .688 & .695 & $+.007$ & .700 & $+.012$ & .639 & .635 & $-.004$ & .631 & $-.008$ \\
CORnet-S V2        & 128  & .757 & .765 & $+.008$ & .758 & $+.001$ & .659 & .649 & $-.010$ & .649 & $-.010$ \\
CORnet-S V4        & 256  & .796 & .799 & $+.003$ & .802 & $+.006$ & .663 & .644 & $-.019$ & .659 & $-.004$ \\
CORnet-S IT        & 512  & .813 & .813 & $+.000$ & .805 & $-.008$ & .666 & .667 & $+.001$ & .664 & $-.002$ \\
CORnet-S Multi     & 960  & .825 & .831 & $+.006$ & .826 & $+.001$ & .670 & .671 & $+.001$ & .670 & $+.000$ \\
VGG-conv3    & 256  & .770 & .766 & $-.004$ & .766 & $-.004$ & .653 & .652 & $-.001$ & .661 & $+.008$ \\
VGG-conv5    & 512  & .752 & .749 & $-.003$ & .751 & $-.001$ & .626 & .630 & $+.004$ & .646 & $+.020$ \\
VGG-conv3 ($4{\times}4$)  & 4096 & .735 & .718 & $-.017$ & .721 & $-.014$ & .669 & .670 & $+.001$ & .670 & $+.001$ \\
VGG-conv5 ($2{\times}2$)  & 2048 & .714 & .710 & $-.004$ & .728 & $+.014$ & .638 & .646 & $+.008$ & .647 & $+.009$ \\
VGG-multi  & 1472 & .811 & .802 & $-.009$ & .806 & $-.005$ & .670 & .669 & $-.001$ & .675 & $+.005$ \\
\midrule
\textbf{Average} &      &      &      & $\mathbf{-.001}$ &      & $\mathbf{+.000}$ &      &      & $\mathbf{-.002}$ &      & $\mathbf{+.002}$ \\
\bottomrule
\end{tabular}
\end{table}

\section{Parameters of Weighted Minkwoski Distance Across the Visual Hierarchy}
\label{app:learned-p}

\begin{table}[t]
\centering
\caption{Learned weighted Minkowski parameters $p$ and per-channel weights $\bm\alpha$ at the saturation point $\nstar$ (in parentheses after $p$). For $\bm\alpha$ we report both the mean and standard deviation across the $\nstar$ learned channels. \underline{Underline} indicates the best-performing model in each case.}
\label{tab:app-learned-p}
\small
\begin{tabular}{@{}lcccc@{}}
  \toprule
       & \multicolumn{2}{c}{BAPPS} & \multicolumn{2}{c}{NIGHTS} \\
  \cmidrule(lr){2-3} \cmidrule(lr){4-5}
  Encoder & $p$ ($\nstar$) & $\bar\alpha\!\pm\!\sigma_\alpha$ & $p$ ($\nstar$) & $\bar\alpha\!\pm\!\sigma_\alpha$ \\
  \midrule
  CORnet-S V1    & 1.01 (4) & .08 $\pm$ .03 & 1.66 (8)  & .67 $\pm$ .03 \\
  CORnet-S V2    & \underline{1.09 (3)} & \underline{.12 $\pm$ .02} & 1.66 (14) & .72 $\pm$ .04 \\
  CORnet-S V4    & 1.28 (4) & .18 $\pm$ .03 & 1.34 (48) & .82 $\pm$ .05 \\
  CORnet-S IT    & 2.29 (6) & .40 $\pm$ .02 & 1.63 (96) & .71 $\pm$ .01 \\
  \midrule
  CORnet-S Multi & 1.36 (3) & .22 $\pm$ .02 & \underline{1.51 (96)} & \underline{.75 $\pm$ .02} \\
  \bottomrule
\end{tabular}
\end{table}

\paragraph{Exponent.}
We use a weighted Minkowski distance (p-norm) to quantify the difference between two images (\Equ{eq:distance}).
In psychophysics, \emph{separable} dimensions---those that observers can attend to and judge independently---yield similarity judgments best described by a city-block metric ($p = 1$), while \emph{integral} dimensions---those perceived as a fused whole---yield Euclidean metrics ($p = 2$)~\citep{nosofsky1986attention,shepard1964attention,garner1974processing}.
Mathematically, at $p = 1$, the relative contribution of a dimension to the distance is independent of the magnitudes of the other dimensions, whereas at $p = 2$, the relative contribution of one dimension depends on the others\footnote{Consider the L2 distance $d = \sqrt{x^2 + y^2}$; $\partial d/\partial x = x/\sqrt{x^2 + y^2}$, which depends on $y$. In contrast, for the L1 distance $d = |x| + |y|$, we have $\partial d/\partial x = \text{sign}(x)$, which is independent of $y$.}.

\Tbl{tab:app-learned-p} shows the Minkowski exponent $p$ that the observer model learns at the saturation point $\nstar$ for each encoder-dataset combination.
On BAPPS, $p$ climbs from values near $1$ at V1 to values near $2$ at IT.
On NIGHTS the same observer settles in a narrower $1.3$--$1.7$ range across the hierarchy.
Looking at the best performing model in each dataset, in the low-level dataset BAPPS, the learn $p$ is much lower than that on the high-level dataset NIGHTS, indicating that low-level IQA is better explained by separable dimensions, whereas high-level IQA is better explained by integral dimensions.

\paragraph{Weights.}
The per-channel weights $\bm\alpha$ in \Equ{eq:distance} compensate for the differing natural magnitudes of perceptual channels.
At the saturation point $\nstar$, the weights are approximately uniform within each encoder: the coefficient of variation $\sigma_\alpha/\bar\alpha$ ranges from $5\%$ to $17\%$ on BAPPS (V1 is the only exception, at $38\%$ with $\nstar = 4$) and is at most $6\%$ on NIGHTS, meaning no channel preferentially dominates the distance.
This is expected, since any difference across dimensions could potentially be learned through the projection matrix itself.

\end{document}